\documentclass[10pt]{article}
\usepackage{fix-cm}
\usepackage{lmodern}
\PassOptionsToPackage{table}{xcolor}
\usepackage{yalenlp}
\usepackage{amsmath,amssymb,amsthm,mathtools}
\usepackage{array,multirow,tabularx}
\usepackage{algorithm,algorithmic}
\usepackage{placeins}
\usepackage{url}
\usepackage{needspace}
\definecolor{methodshade}{gray}{0.94}
\hypersetup{colorlinks=true,allcolors=black}

\newtheorem{theorem}{Theorem}
\newtheorem{proposition}[theorem]{Proposition}
\newtheorem{corollary}[theorem]{Corollary}
\newtheorem{lemma}[theorem]{Lemma}
\theoremstyle{remark}

\newcommand{\KL}{D_{\mathrm{KL}}}
\newcommand{\E}{\mathbb{E}}
\newcommand{\R}{\mathbb{R}}
\newcommand{\one}{\mathbf{1}}
\newcommand{\sg}{\operatorname{sg}}
\newcommand{\softmax}{\operatorname{softmax}}
\newcommand{\clip}{\operatorname{clip}}
\newcommand{\norm}[1]{\left\lVert #1\right\rVert_2}
\newcommand{\method}{OC-OPD}

\title{OC-OPD: Observation-coordinated Supervision for On-Policy Distillation}
\author{%
  \textbf{Ruoran Xu}\qquad\textbf{Yilun Zhao}\qquad
  \textbf{Ziyu Chen}\qquad\textbf{Arman Cohan}\\[7pt]
  Yale NLP Lab
}
\date{}
\hypersetup{pdftitle={OC-OPD: Observation-coordinated Supervision for On-Policy Distillation},
            pdfauthor={Ruoran Xu, Yilun Zhao, Ziyu Chen, Arman Cohan}}

\begin{document}
\maketitle
\thispagestyle{yalefirst}

\begin{abstract}

Reverse and forward Kullback--Leibler (KL) divergences provide complementary supervision in on-policy distillation: reverse KL refines the student's active token rankings, and forward KL allocates mass to teacher-supported candidates. At student-generated prefixes, we form a four-observation matrix by treating the student and teacher as both candidate sources and evaluators. The diagonal self-observations measure uncertainty, while source-centered cross-observations recover the two KL objectives. We establish an exact identity equating the Fisher inverse-metric coupling of their logit gradients with the sum of these contrasts, yielding a common local descent direction for distinct positive distributions. Observation-coordinated on-policy distillation (OC-OPD) uses observations to set preferences, closed-form bargaining to coordinate gradients, and distributional discrepancy to scale the update. Across three heterogeneous student settings (non-thinking generation, thinking generation, and base-model initialization), OC-OPD achieves strong gains in mathematical reasoning and transfers effectively to out-of-distribution scientific question answering and code generation. Our code is publicly available at: \href{https://github.com/RuoranXu/OC-OPD}{https://github.com/RuoranXu/OC-OPD}.

\end{abstract}

\begin{figure}[h]
    \centering
    \includegraphics[width=1\linewidth]{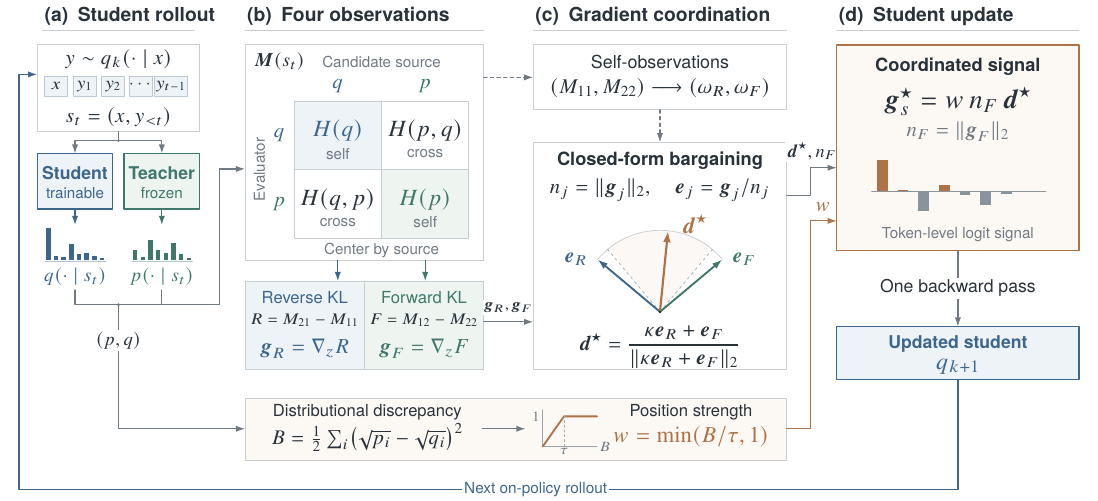}
    \caption{\textbf{Training pipeline of OC-OPD.}
\textbf{(a)} Student rollouts provide prefixes at which the student and frozen teacher produce aligned next-token distributions.
\textbf{(b)} Four self- and cross-observations at the shared prefix yield source-centered KL gradients for ranking/refinement and coverage.
\textbf{(c)} Self-observation values set uncertainty-based priorities for closed-form gradient bargaining, while distributional discrepancy controls supervision strength.
\textbf{(d)} The resulting bounded logit signal updates the student through a single backward pass. The updated student then generates new rollouts.}
    \label{fig:oc_opd_overview}
\end{figure}

\section{Introduction}

On-policy distillation (OPD) trains a student on prefixes it generates, with a frozen teacher supervising the same next-token decision \citep{dagger,gkd}. Reverse KL (RKL) refines how the student allocates probability among the candidates it already explores. Forward KL (FKL) transfers probability toward teacher-supported alternatives, including those the student rarely explores \citep{gkd,minillm}. How should the two corrections contribute to a single update at each student-generated prefix?

Recent OPD methods adapt supervision using uncertainty, token importance, teacher--student mismatch, and divergence routing \citep{reopold,eopd,sead,tip,tide}. We use two properties of the predictions at the same prefix. Each model's entropy describes how broadly it allocates probability, and the disagreement between the two distributions describes what needs correction.

We organize these signals by letting each model evaluate the candidate probabilities supplied by both models. The resulting four-observation matrix records each model's entropy and its cross-entropy on the other model's distribution. Subtracting each source's entropy from the corresponding cross-entropy yields the two KL objectives at the same prefix.

We link the sum of the two KL objectives to the geometry of their logit gradients through a Fisher inverse-metric identity. When the predictions are distinct and positive, the identity establishes that a small step in a common direction can decrease both objectives in student-logit space.

We develop observation-coordinated on-policy distillation (\method{}) around three decisions. The two entropies set the relative \emph{preference} for refinement and coverage. A closed-form bargaining rule selects a \emph{direction} within the common-descent region. Distributional discrepancy sets the \emph{strength} of the update at each prefix. The resulting bounded logit signal is applied with one student backward pass (Figure~\ref{fig:oc_opd_overview}).

We evaluate \method{} on mathematical reasoning in three Qwen3 student settings \citep{qwen}, all trained with fresh student rollouts: non-thinking students at 1.7B and 4B scales, thinking-generation students, and a weaker base-model initialization. In the main non-thinking setting, the 1.7B student exceeds the strongest compared baseline by $+6.88$ Avg@32 and $+11.11$ Pass@32 points on average across AIME24--26; the 4B student exceeds EOPD \citep{eopd} by $+2.74$ and $+3.33$ points, respectively (Table~\ref{tab:main}). Gains persist in the other two settings, and GPQA-Diamond \citep{rein2023gpqa} and MBPP+ \citep{liu2023evalplus} assess transfer to scientific question answering and code generation.

Our contributions are threefold. \textbf{(1)} We organize token-level OPD through a source--evaluator system whose self- and cross-observations characterize the teaching state and whose source-centered contrasts recover the two KL corrections. \textbf{(2)} We prove a Fisher inverse-metric identity linking those contrasts to their logit gradients and establish common local descent for every distinct positive teacher--student pair at a fixed prefix. \textbf{(3)} We develop \method{} to determine the preference, direction, and strength of token-level supervision, and demonstrate its effectiveness across heterogeneous student settings and out-of-domain evaluations.

\section{Related Work}
\label{sec:related}

\paragraph{KL directions in on-policy distillation.}
Earlier knowledge distillation methods transfer predictive structure from a teacher to a student \citep{hinton,seqkd}. AKL adaptively combines forward and reverse KL using distributional gaps within teacher-defined head and tail regions, motivated by their different fitting priorities in early training \citep{akl}. A larger head gap increases the forward-KL weight, while a larger tail gap increases the reverse-KL weight. OPD evaluates teacher guidance at student-generated prefixes \citep{dagger,gkd}, where both models produce aligned next-token distributions. GKD studies flexible divergence objectives at these states \citep{gkd}, and MiniLLM develops reverse-KL distillation under student sampling \citep{minillm}. Related methods explore skewed divergences and adaptive sampling \citep{distillm}, contrastive teacher--student responses \citep{distillm2}, and interleaved generation \citep{skd}. Reverse KL refines relative probabilities among student-supported candidates, while forward KL transfers mass toward teacher-supported candidates. EOPD conditions forward-KL supervision on teacher entropy \citep{eopd}, and SEAD adjusts the emphasis of the two KL directions across tokens \citep{sead}. We study how the current teacher--student state relates to these corrections: source-centering separates self-uncertainty from directed disagreement, and an exact observation--gradient identity establishes their joint local compatibility.

\paragraph{Observation-based adaptive supervision.}
At a shared prefix, either model can supply candidate mass and either model can evaluate it. These choices produce four observations: two self-entropies describe the predictive states of the student and teacher, and two cross-entropies combine source uncertainty with directed cross-model disagreement. Related methods use local information to determine how supervision is assigned. REOPOLD uses entropy-based token selection and relaxed reward shaping \citep{reopold}; SMOPD selects informative positions for self-distillation \citep{smopd}; TIP combines student uncertainty with divergence to assess token importance \citep{tip}, while TIDE characterizes student-excess and student-deficit mismatch \citep{tide}. Region-dependent objectives, adaptive targets, bounded transformations, and trust-region control further shape local supervision \citep{aopd,veto,poweropd,tropd}. Our source--evaluator matrix connects these signals through the entropy decomposition of KL: source-centering isolates the two directed disagreements, whose gradients supply the refinement and coverage corrections. Self-entropies set their relative preferences, gradient geometry coordinates their direction, and distributional discrepancy sets supervision strength. Appendix~\ref{rel} summarizes representative methods in this notation.

\paragraph{Gradient coordination.}
Multi-objective optimization derives shared update directions via minimum-norm descent, gradient projection, conflict-averse optimization, and bargaining. Our KL gradients stem from aligned student–teacher distributions and source-centered observation contrasts, whose local coupling follows an exact Fisher-geometric identity. We leverage this identity to coordinate candidate-level refinement and coverage, adopting bargaining to obtain an observation-conditioned closed-form update direction. Scalarization and normalized gradient mixing serve as baselines for comparison.

\section{Preliminaries and Motivation}
\label{sec:observation}
\label{sec:motivation}
This section formalizes shared-context OPD, organizes the aligned student and teacher distributions into four observations, and shows how their source-centered contrasts recover the ranking and coverage signals of reverse and forward KL.

\subsection{Shared-context OPD as observation system}
OPD evaluates supervision along student-generated responses by querying the teacher at the same prefixes \citep{gkd,minillm}. We distinguish this choice of training states from the information available at each visited state. In the shared-context setting, both models condition on the same prompt and student prefix, yielding an aligned pair of next-token predictive distributions. Our aim is to organize the information in this pair into effective supervision.

For a prompt $x$, let $q_\theta$ and $p_T$ denote the student and frozen teacher. The sequence-level reverse KL is
\begin{equation}
\begin{aligned}
\mathcal J_{\mathrm{OPD}}(\theta;x)
&=\KL\bigl(q_\theta(\cdot\mid x)\Vert p_T(\cdot\mid x)\bigr)\\
&=\E_{y\sim q_\theta(\cdot\mid x)}
\left[\sum_{t=1}^{|y|}\log\frac{q_\theta(y_t\mid s_t)}{p_T(y_t\mid s_t)}\right],
\qquad s_t=(x,y_{<t}).
\end{aligned}
\label{eq:opd_objective}
\end{equation}
Each sampled sum in Eq.~\eqref{eq:opd_objective} estimates the sequence-level KL value. The sequence-level gradient accounts for the future prefixes induced by each sampled token (Appendix~\ref{app:sequence_rkl}).

For token-local updates at iteration $k$, collect valid response positions $\mathcal S_k$ from fresh rollouts of $q_k=q_{\theta_k}$. The surrogate assigns each sampled token a teacher-to-student log-ratio credit.
\begin{equation}
\widehat{\mathcal L}_{\mathrm{OPD},k}(\theta)
=-\frac{1}{N}\sum_{t\in\mathcal S_k}\sg[A_{k,t}]\log q_\theta(y_t\mid s_t),
\qquad A_{k,t}=\log\frac{p_T(y_t\mid s_t)}{q_k(y_t\mid s_t)},
\label{eq:opd_surrogate}
\end{equation}
Here $N=|\mathcal S_k|$ and $\sg$ denotes stop-gradient. With prefixes and credits fixed during differentiation, the sampled-token gradient at $\theta_k$, averaged over $y_t\sim q_k(\cdot\mid s_t)$ at a given $s_t$, gives the conditional reverse KL gradient. We use the aligned next-token distributions to study both conditional KL directions at each collected prefix.

\subsection{Four observations from two distributions}
\label{sec:organization}
At a collected prefix $s=(x,y_{<t})$, write $q=q_{\theta_k}(\cdot\mid s)$ and $p=p_T(\cdot\mid s)$ over an aligned vocabulary. The fixed pair $p$ and $q$ supplies two candidate sources---student and teacher---and two evaluators---student and teacher. Pairing each source with each evaluator yields four expected observations. For $H(a,b)=-\sum_i a_i\log b_i$ and $H(a)=H(a,a)$, define
\begin{equation}
M(s)=\begin{pmatrix}H(q)&H(p,q)\\H(q,p)&H(p)\end{pmatrix}
=-\sum_{i\in\mathcal V}
\begin{pmatrix}q_i\log q_i&p_i\log q_i\\q_i\log p_i&p_i\log p_i\end{pmatrix}.
\label{eq:matrix}
\end{equation}
Rows index evaluators and columns index candidate sources, both ordered as student, teacher. Accordingly, $M_{11}$ records the student evaluating student-sourced mass, $M_{12}$ the student evaluating teacher-sourced mass, $M_{21}$ the teacher evaluating student-sourced mass, and $M_{22}$ the teacher evaluating teacher-sourced mass. The diagonal self-observations are the student and teacher entropies and quantify distributional dispersion. The off-diagonal cross-observations measure how well one model accommodates candidates emphasized by the other.

The matrix provides a relational reparameterization of the same $p,q$ pair by separating the distribution supplying candidate mass from the distribution scoring it. Teacher-sourced expectations use the teacher distribution at the same student-visited prefix. The matrix records state-level expectations, while the underlying $p$ and $q$ retain the candidate-level resolution needed for an update. It gathers common adaptive OPD signals in one object: student entropy is $M_{11}$, teacher entropy is $M_{22}$, and teacher--student mismatch appears as the off-diagonal excess over the corresponding self-observation. Table~\ref{tab:four_obs_comparison} in appendix~\ref{rel} compares how different methods are described in this framework.

\subsection{Source-centered contrasts reveal ranking and coverage}
\label{sec:contrasts}
Centering each column on its candidate source's self-observation isolates the difference between the two evaluators:
\begin{equation}
\begin{aligned}
 R&=M_{21}-M_{11}=H(q,p)-H(q)=\KL(q\Vert p),\\
 F&=M_{12}-M_{22}=H(p,q)-H(p)=\KL(p\Vert q).
\end{aligned}
\label{eq:views}
\end{equation}
Thus reverse and forward KL arise as centered cross-observations from the student-sourced and teacher-sourced columns, respectively.

\paragraph{Student-sourced contrast: rank and refine active support.}
For reverse KL, the token-level credit $A_i=\log(p_i/q_i)$ compares teacher support with the student's current allocation. Since $A_i>A_j$ exactly when $p_i/q_i>p_j/q_j$, this credit ranks student-supported candidates by relative underallocation. The student-sourced contrast thereby refines the alternatives the student already explores. Its logit gradient is weighted by $q_i$, concentrating the correction on candidates that carry student probability.

\paragraph{Teacher-sourced contrast: recover underrepresented support.}
Forward KL evaluates teacher-sourced candidates under the student. Its logit gradient $q-p$ directly reallocates mass toward teacher-supported alternatives, including those with very small student probability. This coverage correction allocates student capacity across the teacher-supported set \citep{minillm}. In our full-vocabulary construction, the teacher-sourced gradient is evaluated directly at the visited prefix, yielding the exact candidate-level correction from the aligned distributions.

\paragraph{From complementary contrasts to a coordination problem.}
The matrix thus provides both statistics describing the current state and contrasts specifying candidate-level corrections. Existing entropy gates, token filters, and mismatch criteria can be interpreted as selecting or weighting particular regions of this object \citep{reopold,eopd,sead,tip,tide}. Together, the source-centered contrasts invite a geometric account of when their ranking/refinement and coverage gradients share a descent direction. Section~\ref{sec:theory} establishes this compatibility; Section~\ref{sec:method} uses self-observation, compatibility, and discrepancy to determine preference, direction, and magnitude.

\section{Theory: From Observations to Compatible Gradients}
\label{sec:theory}
The matrix view yields two source-centered corrections whose candidate-level gradients can be studied together. Their ranking/refinement and coverage directions admit common descent, with the observation matrix providing an explicit certificate of compatibility.

\subsection{From observation entries to candidate-level gradients}
For current-state coefficients $a_s,b_s$, define the local matrix loss
\begin{equation}
\begin{aligned}
\ell_s(\theta)
&=\left\langle\sg\!\left[\begin{pmatrix}-a_s&b_s\\a_s&-b_s\end{pmatrix}\right],M_\theta(s)\right\rangle_F\\
&=\sg[a_s]\bigl[H(q_\theta,p)-H(q_\theta)\bigr]
 +\sg[b_s]\bigl[H(p,q_\theta)-H(p)\bigr].
\end{aligned}
\label{eq:loss}
\end{equation}
Writing the coefficient matrix as $W_s$, its columns preserve each candidate source's self-observation as a baseline. The resulting contrasts are $R=\KL(q\Vert p)$ and $F=\KL(p\Vert q)$; Appendix~\ref{app:observation_gradient} derives these identities from the observations.

Differentiating the matrix with respect to the current student logits yields
\begin{equation}
\begin{aligned}
 g_R&=\nabla_z(M_{21}-M_{11})
     =q\odot\left(\log(q/p)-R\one\right),\\
 g_F&=\nabla_z(M_{12}-M_{22})=q-p,\\
 \nabla_z\ell_s&=-a_s\nabla_zM_{11}+b_s\nabla_zM_{12}+a_s\nabla_zM_{21}-b_s\nabla_zM_{22}
                =a_sg_R+b_sg_F.
\end{aligned}
\label{eq:gradients}
\end{equation}
The student self-observation and both cross-observations supply direct derivatives. The frozen teacher's self-observation has $\nabla_zM_{22}=0$ and conditions the coefficients through its value. Thus all four entries contribute to the update. Section~\ref{sec:method} specifies the coefficients; Appendix~\ref{app:observation_gradient} gives the entrywise derivatives and parameter update.

These derivatives make the complementary roles in Section~\ref{sec:contrasts} explicit. For a fixed positive teacher and $q_j\to0$, their coordinates satisfy $g_{R,j}\to0$ and $g_{F,j}\to-p_j$. The teacher-sourced contrast therefore retains a direct correction for candidates with very small student probability. Coordinating the two directions refines the student's current allocation while preserving a correction toward teacher-favored alternatives.

\subsection{Compatibility certified by the observation contrast}
\label{sec:geometry}
Let $J(q)=\operatorname{diag}(q)-qq^\top$ be the categorical Fisher matrix \citep{amari,nielsen}, with Moore--Penrose pseudoinverse $J(q)^+$. The following identity relates the source-centered gradients through the observation matrix.

\begin{theorem}[Observation--gradient identity]
\label{thm:coupling}
For positive $p,q$, the logit gradients in Eq.~\eqref{eq:gradients} satisfy
\begin{equation}
 g_R^\top J(q)^+g_F
 =\sum_i(q_i-p_i)\log\frac{q_i}{p_i}
 =M_{12}+M_{21}-M_{11}-M_{22}=R+F.
\label{eq:fisher}
\end{equation}
If $p\neq q$, both gradients have positive norm and their Euclidean cosine is strictly greater than $-1$. A unit direction $d$ therefore exists with $g_R^\top d>0$ and $g_F^\top d>0$.
\end{theorem}

For a direction $d$ with both inner products positive, a sufficiently small logit step along $-d$ decreases both conditional KLs at the fixed prefix. The identity guarantees that such directions exist even when the Euclidean gradients form an obtuse angle. Section~\ref{sec:method} selects a direction within this feasible set and scales the update. Appendix~\ref{app:geometry} gives the proof, Appendix~\ref{app:general} extends compatibility to other strictly convex $f$-divergences, and Appendix~\ref{app:update_diagnostics} examines first-order predictions for actual optimizer steps.

\section{Method: Observation-Coordinated Updates}
\label{sec:method}
At each student-visited prefix, the teacher and student provide aligned full-vocabulary distributions. From these we compute the observations in $M(s)$ and the two logit gradients in Eq.~\eqref{eq:gradients}. Algorithm~\ref{alg:ocopd} uses the self-observations to set preferences, selects a direction from the common-descent region of Theorem~\ref{thm:coupling}, and scales it by the disagreement at the prefix. The resulting signal is held fixed during the student backward pass.

\begin{algorithm}[ht]
\caption{Observation-Coordinated On-Policy Distillation}
\label{alg:ocopd}
\small
\setboolean{ALC@noend}{true}
\algsetup{indent=1.2em}
\begin{algorithmic}[1]
\setlength{\itemsep}{0pt}
\setlength{\parsep}{0pt}

\REQUIRE Frozen teacher $p_T$, student $q_\theta$
\REQUIRE Entropy scale $h_0$, preference floor $\epsilon$
\REQUIRE Discrepancy scale $\tau$, learning rate $\eta$
\REQUIRE Training dataset $\mathcal D$, prompt batch size $m$

\FOR{each training iteration $k$}
    \STATE Set $q_k\leftarrow q_{\theta_k}$ and
    $\mathcal R\leftarrow\varnothing$, then
    sample $m$ prompts $\mathcal B\subset\mathcal D$.

    \FOR{each prompt $x\in\mathcal B$}
        \STATE Sample $\mathbf y=(y_1,\ldots,y_{|\mathbf y|})
        \sim q_k(\cdot\mid x)$.

        \FOR{$t=1$ \TO $|\mathbf y|$}
            \STATE $s_t\leftarrow(x,y_{<t})$,\quad
            $q\leftarrow q_k(\cdot\mid s_t)$,\quad
            $p\leftarrow p_T(\cdot\mid s_t)$.

            \STATE $M\leftarrow
            \left[\begin{smallmatrix}
                H(q) & H(p,q)\\
                H(q,p) & H(p)
            \end{smallmatrix}\right]$,\quad
            $R\leftarrow M_{21}-M_{11}$,\quad
            $F\leftarrow M_{12}-M_{22}$.

            \STATE $g_{s_t}^{\star}\leftarrow 0$.
            \IF{$p\neq q$}
                \STATE $g_R\leftarrow
                q\odot\bigl(\log(q/p)-R\mathbf 1\bigr)$,\quad
                $g_F\leftarrow q-p$.

                \STATE Compute $(\omega_R,\omega_F)$ from
                $(M_{11},M_{22})$ by Eq.~\eqref{eq:main_preferences}.

                \STATE Compute $d^\star$ from $g_R,g_F,\omega_R,\omega_F$
                by Eq.~\eqref{eq:direction_closed}.

                \STATE $w\leftarrow\min\!\left\{
                \frac{1}{2\tau}\sum_i
                (\sqrt{p_i}-\sqrt{q_i})^2,\,1\right\}$,\quad
                $g_{s_t}^{\star}\leftarrow
                w\lVert g_F\rVert_2d^\star$.
            \ENDIF
        \ENDFOR

        \STATE Store $(x,\mathbf y,
        \{g_{s_t}^{\star}\}_{t=1}^{|\mathbf y|})$ in $\mathcal R$.
    \ENDFOR

    \STATE Collect valid prefixes $\mathcal S_k$ from $\mathcal R$.

    \STATE $\displaystyle
    L_k(\theta)\leftarrow\frac{1}{|\mathcal S_k|}
    \sum_{s\in\mathcal S_k}\sum_i
    \mathrm{sg}[g_{s,i}^{\star}]\log q_\theta(i\mid s)$.

    \STATE Update student to $\theta_{k+1}$ with one backward pass
    on $L_k$ at learning rate $\eta$.
\ENDFOR

\end{algorithmic}
\end{algorithm}

\subsection{Observation-conditioned gradient coordination}
\label{sec:direction}
A diffuse teacher spreads probability mass across several candidates, motivating a higher preference for coverage. When the teacher is concentrated and the student is diffuse, refinement gives more weight to the teacher's evaluation of the candidates the student explores. We map the two entropies to $[0,1)$ with scale $h_0>0$ and use a floor $\epsilon>0$ to keep both preferences positive.
\begin{equation}
\begin{aligned}
 h_S&=\frac{M_{11}}{M_{11}+h_0},\qquad
 h_T=\frac{M_{22}}{M_{22}+h_0},\\
 \nu_R&=\epsilon+(1-h_T)h_S,\qquad \nu_F=\epsilon+h_T,
 \qquad \omega_j=\frac{\nu_j}{\nu_R+\nu_F},\quad j\in\{R,F\}.
\end{aligned}
\label{eq:main_preferences}
\end{equation}
For $p\neq q$, set $n_j=\norm{g_j}$, $e_j=g_j/n_j$, and $c=e_R^\top e_F$. The quantity $e_j^\top d$ is the first-order gain along $-d$ normalized by $n_j$. When $c<0$, a direct mixture $v=\omega_R e_R+\omega_F e_F$ can give $e_F^\top v=\omega_F+c\omega_R\leq 0$. We use gradient bargaining \citep{nashmtl} to allocate positive gains under the preferences $\omega_j$.
\begin{equation}
 d^*=\arg\max_{\substack{\norm d\leq1\\e_R^\top d>0,\ e_F^\top d>0}}
 \left\{\omega_R\log(e_R^\top d)+\omega_F\log(e_F^\top d)\right\}.
\label{eq:nash}
\end{equation}
Theorem~\ref{thm:coupling} guarantees a nonempty feasible set. The logarithm gives smaller gains larger marginal weight at a fixed preference. The optimization problem in Eq.~\eqref{eq:nash} has the unique solution
\begin{equation}
\begin{aligned}
 \kappa&=\frac{c(\omega_R-\omega_F)
 +\sqrt{c^2(\omega_R-\omega_F)^2+4\omega_R\omega_F}}{2\omega_F},\\
 d^*&=\frac{\kappa e_R+e_F}{D},
 \qquad D=\sqrt{\kappa^2+2\kappa c+1}.
\end{aligned}
\label{eq:direction_closed}
\end{equation}
Equal preferences yield the common-descent direction $(e_R+e_F)/\norm{e_R+e_F}$. Appendix~\ref{app:nash} proves the closed form and gives its stable evaluation. Section~\ref{sec:ablation_study} compares bargaining with direct mixing and with equal preferences.

\subsection{Position strength and realized observation coefficients}
\label{sec:strength}
\label{sec:properties}
The unit vector $d^*$ sets the direction. We use the forward-KL gradient norm $n_F=\norm{q-p}$ as a reference magnitude and attenuate the signal where the distributions nearly agree. The squared Hellinger distance $B=\tfrac12\sum_i(\sqrt{p_i}-\sqrt{q_i})^2$ gives a symmetric discrepancy in $[0,1]$. For $\tau>0$, the gate $w=\min(B/\tau,1)$ increases linearly up to $B=\tau$ and saturates thereafter.
\begin{equation}
\begin{aligned}
 g_s^*&=wn_Fd^*=a_sg_R+b_sg_F,\\
 a_s&=\frac{wn_F\kappa}{Dn_R},\qquad b_s=\frac{w}{D}.
\end{aligned}
\label{eq:update}
\end{equation}
The realized coefficients $a_s,b_s$ instantiate $W_s$ in Eq.~\eqref{eq:loss} and remain fixed during differentiation. When $p=q$, Algorithm~\ref{alg:ocopd} sets $g_s^*=0$ before gradient normalization.

\paragraph{Update properties.}
The signal obeys $\norm{g_s^*}\leq\sqrt2$. For $q_j<p_j$, its correction satisfies $-g_{s,j}^*\geq w\omega_F(p_j-q_j)$, retaining a positive limiting bound as $q_j\to0$ for a fixed interior teacher. Near agreement with an interior teacher, discrepancy gating gives cubic attenuation in $\norm{q-p}$. When the gradients align, the signal reduces to $w(q-p)$ (Appendices~\ref{app:guarantees}, \ref{app:asymptotic}, and~\ref{app:reductions}).

Since $g_s^*$ sums to zero, the positive-sign surrogate
$\widetilde{\mathcal L}_s=\sum_i\sg[g_{s,i}^*]\log q_\theta(i\mid s)$
has logit gradient $g_s^*$ (Eq.~\eqref{eq:surrogate}). The corresponding direct logit descent step is $-\eta g_s^*$. Algorithm~\ref{alg:ocopd} aggregates the surrogate over valid response positions and backpropagates once through the student. Appendix~\ref{app:implementation} details the parameter update.

\FloatBarrier

\section{Experiments}
\label{sec:experiments}
\label{sec:lmeval}

\subsection{Experimental settings}
\paragraph{Models and training data.}
Students are Qwen3-1.7B and Qwen3-4B, with Qwen3-4B-Base used for the base-initialization study \citep{qwen}. We use \emph{Qwen3-30B-A3B-Instruct-2507} and \emph{Qwen3-30B-A3B-Thinking-2507} as teachers \citep{qweninstruct2507,qwenthinking2507}. The OLMo comparison in Appendix~\ref{app:second-model-family} pairs the SFT-initialized Olmo-3-7B-Instruct student with the Olmo-3.1-32B-Instruct teacher \citep{olmo2026olmo3}. The main setting pairs the Instruct teacher with non-thinking students; auxiliary settings use the Thinking teacher with thinking students and the Instruct teacher with Qwen3-4B-Base. All methods train on DAPO-Math-17k \citep{dapo} with fresh student rollouts; distillation methods query a frozen teacher at the same student prefixes. For OPD and OC-OPD, best checkpoint selection uses a held-out DAPO-Math-17k validation set separate from training and the reported test benchmarks (Appendix~\ref{app:protocol}).

\paragraph{Evaluation benchmarks and metrics.}
We evaluate AIME24--26, MATH500 \citep{lightman2024lets}, and OlympiadBench \citep{he2024olympiadbench} for mathematical reasoning and GPQA-Diamond \citep{rein2023gpqa} and MBPP+ \citep{liu2023evalplus} for out-of-distribution transfer. Avg@32 is the mean accuracy over 32 responses per problem and Pass@32 the fraction of problems solved by at least one; OOD columns report Avg@32 in Table~\ref{tab:main} and Pass@32 in Table~\ref{tab:main_results}. AIME24 and AIME25 results use repeated decoding evaluations of fixed checkpoints. Appendix~\ref{app:aime_stats} describes the six decoding repetitions. The tables indicate where Avg@8 or Pass@8 is used. We compare with GRPO \citep{shao2024deepseekmath}, OPD \citep{agarwal2024onpolicy}, OPSD \citep{zhao2026selfdistilled}, EOPD \citep{eopd}, and KD \citep{hinton,seqkd}, as applicable. The 4B non-thinking setting also reports pure RKL and FKL objectives. Within each comparison, methods share initialization, prompts, rollout budget, decoding, and evaluation protocol, and teacher-based methods share the same teacher. Appendix~\ref{computational} reports full-run timing and per-update profiling.


\subsection{Non-thinking student results}
\label{sec:math_results}
\label{sec:main_experiment}
\begin{table}[htbp]
\centering
\caption{\textbf{Results for non-thinking students} (accuracy \%). Shaded rows denote \method{}. AIME24 and AIME25 use repeated decoding evaluations. Dashes indicate unavailable scores. $\Delta$ reports the Qwen3-4B gain over EOPD in percentage points, computed from the displayed scores.}
\label{tab:main}

\fontsize{8}{9.5}\selectfont
\setlength{\tabcolsep}{2.2pt}
\renewcommand{\arraystretch}{1.0}

\resizebox{\linewidth}{!}{%
\begin{tabular}{l*{10}{c}}
\toprule

\multirow{3}{*}{\textbf{Methods}}
& \multicolumn{8}{c}{\textbf{Math (I.D.)}}
& \multicolumn{2}{c}{\textbf{O.O.D.}} \\

\cmidrule(lr){2-9}
\cmidrule(lr){10-11}

& \multicolumn{2}{c}{\textbf{AIME 2024}}
& \multicolumn{2}{c}{\textbf{AIME 2025}}
& \multicolumn{2}{c}{\textbf{AIME 2026}}
& \textbf{MATH500}
& \textbf{OlympiadBench}
& \multicolumn{1}{c}{\textbf{GPQA$_D$}}
& \multicolumn{1}{c}{\textbf{MBPP$+$}} \\

\cmidrule(lr){2-3}
\cmidrule(lr){4-5}
\cmidrule(lr){6-7}

& Avg@32 & Pass@32
& Avg@32 & Pass@32
& Avg@32 & Pass@32
& Avg@8
& Pass@8
& \multicolumn{2}{c}{Avg@32} \\

\midrule

\multicolumn{11}{l}{
    \emph{Teacher: Qwen3-30B-A3B-Instruct-2507}
} \\

Qwen3-1.7B
& 14.16 & 43.33
& 9.90  & 30.00
& 7.71  & 23.33
& 71.88 & 52.67 & 29.59  &  56.03     \\

\ \ + GRPO
& 34.17 & 70.00
& 25.52 & 46.67
& 26.04 & 50.00
& 82.21 & 62.57 &  ---   &    ---   \\

\ \ + OPD
& 31.98 & 73.33
& 20.73 & 50.00
& 23.54 & 46.67
& 85.84 & 60.77 &    ---   &   ---    \\

\ \ + OPSD
& 33.33 & 73.33
& 22.60 & 53.33
& 21.35 & 56.67
& 86.20 & 64.84 &  ---    &  ---   \\

\rowcolor{methodshade}
\ \ + \method{}
& 41.67 & 76.67
& 34.38 & 63.33
& 30.31 & 76.67
& 87.40 & 67.40 & 34.01 &  59.01\\

\cmidrule(lr){1-11}

Qwen3-4B
& 23.85 & 56.67
& 22.40 & 56.67
& 15.10 & 53.33
& 79.20 & 69.73 & 39.60 & 65.55  \\

\ \ + OPD (RKL)
& 61.04 & 86.67
& 51.04 & 76.67
& 52.19 & 80.00
& 87.40 & 73.04 & 43.43 &  64.29 \\

\ \ + OPD (FKL)
& 55.42 & 83.33
& 53.75 & 73.33
& 54.17 & 83.33
& 85.20 & 75.65 & 40.91 & 65.55\\

\ \ + EOPD
& 60.10 & 86.67
& 53.13 & 83.33
& 54.69 & 83.33
& 92.80 & 78.93 & 45.45 & 64.81\\

\rowcolor{methodshade}
\ \ + \method{}
& 60.94 & 93.33
& 56.67 & 83.33
& 58.54 & 86.67
& 94.43 & 80.86 & 46.80 & 66.01  \\

\ \ $\Delta$ vs. EOPD
& +0.84 & +6.66
& +3.54 & +0.00
& +3.85 & +3.34
& +1.63 & +1.93 & +1.35 & +1.20\\

\bottomrule
\end{tabular}%
}
\end{table}

\paragraph{Performance on mathematical reasoning.}
Table~\ref{tab:main} shows consistent gains for both non-thinking student scales. For Qwen3-1.7B, \method{} outperforms GRPO, OPD, and OPSD on all six AIME metrics, beating the strongest baseline by +6.88 Avg@32 and +11.11 Pass@32 points on average. For Qwen3-4B, it exceeds EOPD by +2.74 Avg@32 and +3.33 Pass@32 points across AIME24--26. \method{} is best or tied-best in five of six columns, with RKL higher only on AIME24 Avg@32 by 0.10 points. Thus, the gains cover both average accuracy and multi-sample success.

\paragraph{Combining the two KL directions.}
Neither KL direction dominates at 4B: RKL leads FKL by 5.62 Avg@32 points on AIME24, whereas FKL is stronger on AIME25--26 and keeps MBPP+ at initialization. \method{} retains both strengths: it is within 0.10 points of RKL on AIME24, exceeds the stronger one by +2.92 and +4.37 Avg@32 points on AIME25--26, and surpasses both OOD. It also beats the adaptive linear mixture on all six AIME metrics (Table~\ref{tab:math_ablations}), so the gain comes from coordination rather than mixing alone.

\subsection{Out-of-distribution evaluation}
\label{sec:ood_results}
\paragraph{Generalization beyond mathematical reasoning.}
Although trained only on DAPO-Math-17k, \method{} also improves out-of-distribution performance (Table~\ref{tab:main}, last two columns). Relative to its initialization, the 1.7B student gains +4.42 points on GPQA-Diamond and +2.98 points on MBPP+. At 4B, it reaches 46.80 and 66.01, exceeding EOPD by +1.35 and +1.20 points. On MBPP+, \method{} stays above initialization, whereas EOPD and OPD (RKL) drop by 0.74 and 1.26 points.

\subsection{Base-model initialization}
\label{sec:base_study}

\begin{table*}[ht]
\centering
\begin{minipage}[t]{0.9\textwidth}
\centering
\caption{Avg@8 and Pass@8 for the Qwen3-4B-Base student. AIME24 and AIME25 use repeated decoding evaluations. Shaded rows denote \method{}. Bold denotes the best value in each column.}
\label{tab:base}

\fontsize{7.5}{9}\selectfont
\setlength{\tabcolsep}{2.8pt}
\renewcommand{\arraystretch}{0.96}
\resizebox{\linewidth}{!}{%
\begin{tabular}{l*{7}{c}}
\toprule
\textbf{Methods}
& \multicolumn{2}{c}{\textbf{AIME 2024}}
& \multicolumn{2}{c}{\textbf{AIME 2025}}
& \textbf{OlympiadBench}
& \multicolumn{2}{c}{\textbf{MATH500}} \\
\cmidrule(lr){2-3}
\cmidrule(lr){4-5}
\cmidrule(lr){7-8}
& Avg@8 & Pass@8
& Avg@8 & Pass@8
& Pass@8
& Avg@8 & Pass@8 \\
\midrule

Qwen3-4B-Base & 12.19 & 23.33 & 7.29 & 23.33 & 52.52 & 48.60 & 85.20 \\
\ \ + KD   & 12.50 & 26.67 & 12.08 & 23.33 & 62.43 & 74.73 & 92.00 \\
\ \ + OPD  & 18.91 & 33.33 & 12.50 & 30.00 & 58.80 & 78.40 & 90.40 \\
\ \ + EOPD & 17.92 & 33.33 & 17.50 & 33.33 & 62.61 & 80.05 & 92.60 \\
\rowcolor{methodshade}
\ \ + \method{} & \textbf{21.67} & \textbf{40.00} & \textbf{18.75} & \textbf{40.00} & \textbf{63.20} & \textbf{81.75} & \textbf{93.20} \\

\bottomrule
\end{tabular}%
}
\end{minipage}
\end{table*}

\begin{figure*}[ht]
\centering
\begin{minipage}[t]{0.49\textwidth}
\centering
\includegraphics[width=\linewidth]{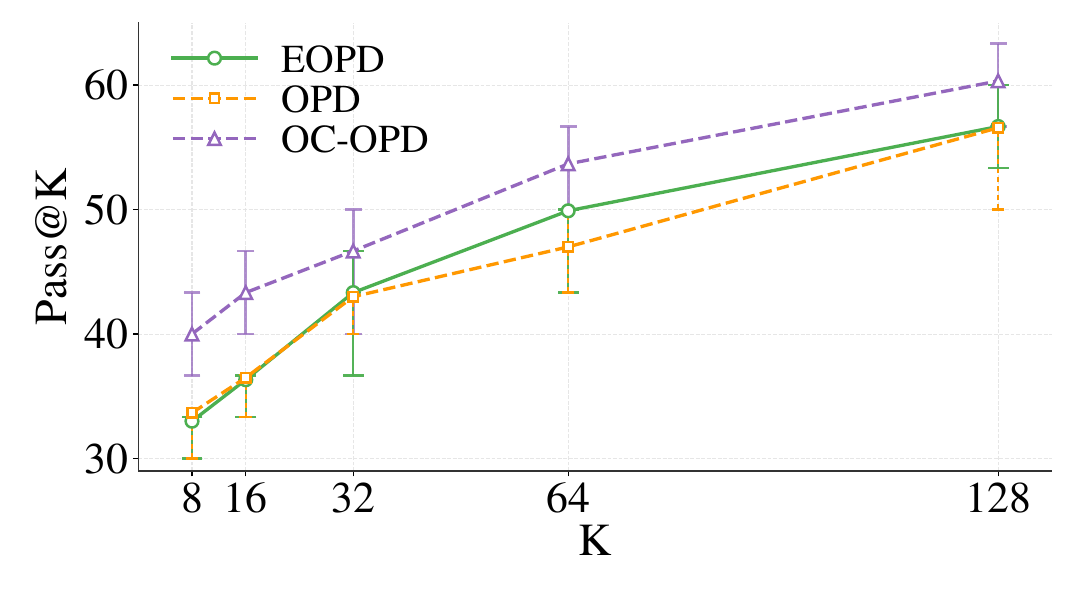}
\captionsetup{skip=4pt,justification=raggedright,singlelinecheck=false}
\captionof{figure}{Pass@k on AIME24 from repeated decoding evaluations with the Qwen3-4B-Base student across sampling budgets.}
\label{fig:passk}
\end{minipage}
\hfill
\begin{minipage}[t]{0.49\textwidth}
\centering
\includegraphics[width=\linewidth]{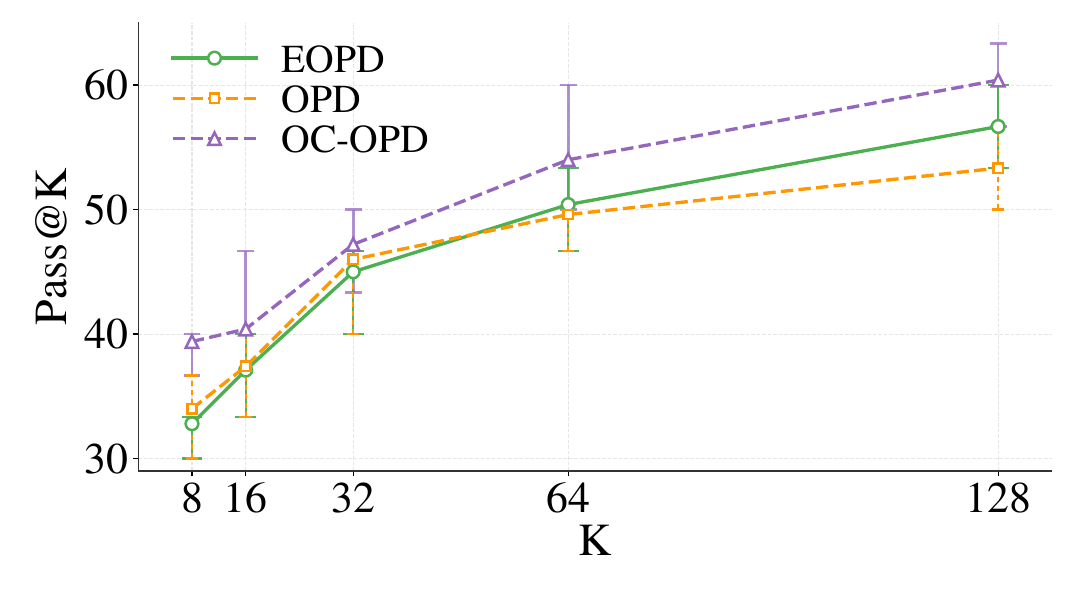}
\captionsetup{skip=4pt,justification=raggedright,singlelinecheck=false}
\captionof{figure}{Pass@k on AIME25 from repeated decoding evaluations with the Qwen3-4B-Base student across sampling budgets.}
\label{fig:passk-right}
\end{minipage}
\end{figure*}


\paragraph{Performance from base-model initialization.}
We further test whether the gains depend on instruction-tuned initialization by distilling into Qwen3-4B-Base. Table~\ref{tab:base} shows that \method{} reaches 40.00 Pass@8 on both AIME24 and AIME25, improving the base model by +16.67 points and EOPD by +6.67 points on each benchmark. Figures~\ref{fig:passk} and~\ref{fig:passk-right} show the Pass@k point estimates from the sampling-budget sweeps. The markers show the reported modal Pass@K for each setting. This is the most frequently observed solved-problem count across the six decoding evaluations, expressed as a percentage. The whisker endpoints of the error bars correspond to the recorded upper and lower bounds.


\subsection{Thinking teacher and thinking students}
\label{sec:thinking_study}
\label{sec:capacity_study}
\paragraph{Performance with thinking generation.}
Table~\ref{tab:main_results} shows that the advantage of \method{} persists with thinking generation. Relative to OPD, the 1.7B student improves by +5.76 Avg@32 and +2.22 Pass@32 points on average across AIME24--26, while the 4B student gains +1.91 and +7.78 points, respectively. \method{} also exceeds OPD on both GPQA-Diamond and MBPP+ at both model scales. These results show that the proposed update is not specific to non-thinking decoding and remains effective when the student already produces explicit reasoning traces.

\begin{table}[ht]
\centering
\begin{minipage}[t]{1\textwidth}
\caption{\textbf{Thinking-model results} (accuracy \%). Shaded rows denote \method{}. AIME24 and AIME25 use repeated decoding evaluations.}
\label{tab:main_results}

\small
\setlength{\tabcolsep}{3.6pt}
\renewcommand{\arraystretch}{1.05}

\resizebox{\linewidth}{!}{%
\begin{tabular}{l*{10}{c}}
\toprule

\multirow{3}{*}{\textbf{Methods}}
& \multicolumn{8}{c}{\textbf{Math (I.D.)}}
& \multicolumn{2}{c}{\textbf{O.O.D.}} \\

\cmidrule(lr){2-9}
\cmidrule(lr){10-11}

& \multicolumn{2}{c}{\textbf{AIME24}}
& \multicolumn{2}{c}{\textbf{AIME25}}
& \multicolumn{2}{c}{\textbf{AIME26}}
& \multicolumn{2}{c}{\textbf{OlympiadBench}}
& \multicolumn{1}{c}{\textbf{GPQA$_D$}}
& \multicolumn{1}{c}{\textbf{MBPP$+$}} \\

& avg@32 & pass@32
& avg@32 & pass@32
& avg@32 & pass@32
& avg@8 & pass@8
& \multicolumn{2}{c}{pass@32} \\

\midrule

1.7B
& 50.94 & 83.33
& 37.19 & 73.33
& 38.65 & 70.00
& 59.33 & 79.38
& 77.78 & 84.66 \\

\ \ + OPD
& 48.54 & 86.67
& 40.21 & 73.33
& 42.08 & 73.33
& 58.91 & 79.08
& 79.29 & 83.60 \\

\rowcolor{methodshade}[\tabcolsep][\tabcolsep]
\ \ + \method{}
& 52.81 & 86.67
& 46.46 & 76.66
& 48.85 & 76.67
& 59.70 & 80.12
& 81.82 & 85.45\\

4B
& 72.71 & 86.67
& 65.52 & 83.33
& 61.04 & 83.33
& 69.08 & 84.91
& 84.34 & 87.04 \\

\ \ + OPD
& 78.13 & 83.33
& 65.94 & 80.00
& 64.17 & 83.33
& 69.88 & 85.72
& 81.31 & 87.57 \\

\rowcolor{methodshade}[\tabcolsep][\tabcolsep]
\ \ + \method{}
& 80.21 & 93.33
& 66.98 & 86.67
& 66.77 & 90.00
& 71.90 & 88.09
& 85.35 & 88.36\\

\bottomrule
\end{tabular}%
}
\end{minipage}
\end{table}

\begin{table}[htbp]
\centering
\caption{\textbf{Ablation study.} The student is Qwen3-4B with non-thinking generation and the teacher is Qwen3-30B-A3B-Instruct-2507. AIME24 and AIME25 use repeated decoding evaluations. Bold denotes the best value in each column.}
\label{tab:math_ablations}
\small
\setlength{\tabcolsep}{5.0pt}
\renewcommand{\arraystretch}{1.05}
\begin{tabularx}{\linewidth}{>{\raggedright\arraybackslash}p{0.25\linewidth}*{6}{>{\centering\arraybackslash}X}}
\toprule
\multirow{2}{*}{\textbf{Method}}
& \multicolumn{2}{c}{\textbf{AIME 2024}}
& \multicolumn{2}{c}{\textbf{AIME 2025}}
& \multicolumn{2}{c}{\textbf{AIME 2026}} \\
\cmidrule(lr){2-3}\cmidrule(lr){4-5}\cmidrule(lr){6-7}
& Avg@32 & Pass@32 & Avg@32 & Pass@32 & Avg@32 & Pass@32 \\
\midrule
\rowcolor{methodshade}
\method{}
& 60.94 & \textbf{93.33}
& \textbf{56.67} & \textbf{83.33}
& \textbf{58.54} & \textbf{86.67} \\
Forward KL (FKL)
& 55.42 & 83.33
& 53.75 & 73.33
& 54.17 & 83.33 \\

Reverse KL (RKL)
& \textbf{61.04} & 86.67
& 51.04 & 76.67
& 52.19 & 80.00 \\

Adaptive linear mixture
& 60.21 & 86.67
& 53.02 & 80.00
& 53.96 & 80.00 \\

Equal-preference
& 58.65 & 86.67
& 53.54 & 80.00
& 56.67 & 83.33 \\

w/o position gate
& 60.63 & 90.00
& 54.06 & 80.00
& 56.67 & 83.33 \\
\bottomrule
\end{tabularx}
\end{table}

\FloatBarrier

\subsection{Ablation study}
\label{sec:ablation_study}
\paragraph{Effect of individual design choices.}
Table~\ref{tab:math_ablations} compares three components of the update. The equal-preference variant sets both preferences to $1/2$ and retains the bargaining objective and position gate. The adaptive linear mixture uses a normalized weighted sum of the gradients with the same entropy-based preferences and magnitude rule. \method{} improves Avg@32 and Pass@32 over both variants on all three AIME editions. Setting $w=1$ retains the direction rule and removes the discrepancy-dependent scaling. With the gate, \method{} gains about 3.33 Pass@32 points on each edition. Appendix~\ref{app:ablations} gives the control definitions, and Appendix~\ref{app:aime_stats} describes the repeated decoding evaluations.


\section{Conclusion}
\label{sec:discussion}
OC-OPD organizes teacher and student predictions at each student-generated prefix into four self- and cross-observations, yielding a unified supervision state. Source-centered contrasts recover the complementary ranking and coverage corrections of reverse and forward KL. The observation--gradient identity establishes their local compatibility, while observation-conditioned coordination combines them in a bounded update that retains a correction for teacher-supported candidates with low student probability. Experiments across non-thinking, thinking, and base-initialized students show consistent gains in mathematical reasoning. Scientific question answering and code generation further assess transfer beyond the mathematical training domain.
\label{mainend}

\clearpage
\bibliographystyle{unsrtnat}
\bibliography{references}
\clearpage
\appendix
\section{Notation, assumptions, and the observation matrix}
\label{app:notation}
\begin{table}[ht]
\centering
\caption{Notation used throughout the paper. All local quantities refer to the same current prefix.}
\begin{tabular}{p{0.20\linewidth}p{0.73\linewidth}}
\toprule
Symbol & Meaning\\
\midrule
$s=(x,y_{<t})$ & A prefix visited by the current student behavior policy\\
$p,q\in\Delta_V^\circ$ & Frozen teacher and current student distributions with positive support\\
$z\in\R^V$ & Student logits, with $q=\softmax(z)$\\
$M$ & Self-/cross-entropy observation matrix in Eq.~\eqref{eq:matrix}\\
$R,F$ & Student-sourced and teacher-sourced observation contrasts\\
$g_R,g_F$ & Their Euclidean gradients with respect to current logits\\
$J(q)$ & $\operatorname{diag}(q)-qq^\top$, with kernel $\operatorname{span}\{\one\}$\\
$n_j,e_j$ & Gradient norm $\norm{g_j}$ and normalized gradient $g_j/n_j$\\
$c$ & Euclidean cosine $e_R^\top e_F$\\
$\omega_R,\omega_F$ & Positive bargaining preferences summing to one\\
$d^*$ & Unit-norm bargaining direction, used with a minus sign in descent\\
$B,w$ & Squared Hellinger discrepancy and position weight\\
$g^*$ & Final logit signal $wn_Fd^*$\\
$\sg[\cdot]$ & Stop-gradient operator evaluated at the current snapshot\\
$A_s,K_{st}$ & Parameter-to-logit Jacobian and block coupling $A_sA_t^\top$\\
\bottomrule
\end{tabular}
\end{table}

\paragraph{Interior support.}
The theoretical domain is $\Delta_V^\circ=\{q\in\R^V:q_i>0,\sum_iq_i=1\}$. Theorems~\ref{thm:coupling} and~\ref{thm:nash} hold throughout this open simplex, including arbitrarily small positive coordinates. The expansion around agreement and the coordinate limit $g_{R,j}\to0$ use a fixed interior teacher. Boundary limits are taken through interior distributions, with divergence values interpreted in the extended-real sense at boundary points.

\paragraph{Freezing conventions.}
The teacher probabilities, prefix samples, preferences, gate, and bargaining solution are detached during the local optimization step. The student probabilities inside $R,F$ in Eq.~\eqref{eq:loss}, or inside $\log q_\theta$ in Eq.~\eqref{eq:surrogate}, remain differentiable. The differentiated objective is conditional on the current prefix distribution. Fresh student sampling supplies the state distribution for each subsequent update.

\paragraph{Sequence-level reverse KL gradient.}
\label{app:sequence_rkl}
For a fixed prompt $x$, write $c_t=\log(q_\theta(y_t\mid s_t)/p_T(y_t\mid s_t))$ and $u_t=\nabla_\theta\log q_\theta(y_t\mid s_t)$. Differentiating Eq.~\eqref{eq:opd_objective} gives
\[
\nabla_\theta\mathcal J_{\mathrm{OPD}}(\theta;x)
=\E_{y\sim q_\theta(\cdot\mid x)}\left[\sum_{t=1}^{|y|}u_t\sum_{v=t}^{|y|}c_v\right].
\]
The derivative of the sampling distribution yields $(\sum_t u_t)(\sum_v c_v)$, and differentiating the log ratios yields $\sum_t u_t$. Since $\E[u_t\mid s_t]=0$, the latter term and all products $u_tc_v$ with $v<t$ vanish in expectation. At a fixed prefix, the same score calculation gives $\E[u_tc_t\mid s_t]$ for the conditional reverse KL gradient, which the token term in Eq.~\eqref{eq:opd_surrogate} estimates at $\theta_k$ before batch averaging.

\paragraph{What each observation contributes.}
$M_{11}$ supplies student dispersion and the student-source baseline. $M_{22}$ supplies teacher dispersion and the teacher-source baseline. The cross-observations $M_{21}$ and $M_{12}$ evaluate each source under the other model. Their within-source differences define centered supervision signals, and their total excess supplies the geometric certificate. Frozen teacher entropy influences the adaptive coefficients through its numerical value.

\paragraph{Organized observations and candidate resolution.}
The matrix provides a canonical organization of four expected scores. The aligned distributions $p,q$ provide candidate-level probabilities, likelihood ratios, and gradient geometry. \method{} uses this information jointly: $w$ determines position strength and $g^*$ allocates the learning signal. Every candidate remains represented in the gradient construction.

\subsection{From the four observations to a parameter gradient}
\label{app:observation_gradient}
The convention $H(a,b)=-\sum_i a_i\log b_i$ averages the evaluator's negative log probability under candidate source $a$. Each column fixes that source while changing the evaluator. The centered contrasts in Eq.~\eqref{eq:views} therefore admit the representations
\[
\begin{aligned}
 R&=H(q,p)-H(q)=\sum_iq_i\log(q_i/p_i)=\KL(q\Vert p),\\
 F&=H(p,q)-H(p)=\sum_ip_i\log(p_i/q_i)=\KL(p\Vert q).
\end{aligned}
\]
These identities express the source-centered observation costs in standard divergence notation.

\paragraph{Entrywise derivatives.}
At a fixed prefix, let $q=q_\theta=\softmax(z)$ and hold $p$ fixed. Direct differentiation of the four entries gives
\begin{equation}
\begin{aligned}
\nabla_z M_{11}&=-q\odot\big(\log q+H(q)\one\big),\\
\nabla_z M_{12}&=q-p,\\
\nabla_z M_{21}&=-q\odot\big(\log p+H(q,p)\one\big),\\
\nabla_z M_{22}&=0.
\end{aligned}
\label{eq:entrywise}
\end{equation}
For the first identity, $\nabla_q H(q)=-(\log q+\one)$ and multiplication by $J(q)$ subtracts the $q$-mean. For the third, $\nabla_q H(q,p)=-\log p$, so the same centering gives the stated expression. The second is the usual cross-entropy logit derivative, and the last follows because the teacher is frozen.

Substituting these derivatives into the matrix loss yields
\begin{equation}
\begin{aligned}
\nabla_z\ell_s
 &=-a_s\nabla_zM_{11}+b_s\nabla_zM_{12}
   +a_s\nabla_zM_{21}-b_s\nabla_zM_{22}\\
 &=a_s q\odot\big(\log(q/p)-R\one\big)+b_s(q-p).
\end{aligned}
\label{eq:matrix_gradient}
\end{equation}
The student self-observation contributes the centering term through its derivative. Combining it with the teacher's evaluation gives the student-sourced direction $g_R$. For the teacher-sourced contrast, the constant $H(p)$ establishes the zero-at-agreement baseline, and $H(p,q)$ supplies the derivative $g_F$.

\paragraph{Values condition the update; derivatives provide its direction.}
All four entries are evaluated at the current snapshot. The diagonal values condition the preference rule in Eq.~\eqref{eq:preferences}; the two excesses and their derivatives define the candidate-level objectives and geometry. Teacher entropy influences the numerical coefficients through the conditioning path, with a zero direct derivative. Alignment and gate saturation can make some coefficient sensitivities vanish at particular pairs, while the observation-to-gradient map retains all four inputs.

The coefficient structure $W_s$ pairs each source's self- and cross-observation through one realized coefficient. The frozen-coefficient convention differentiates $M_\theta$ with $a_s,b_s$ held at their current values. In the product-rule expansion, the derivative of these fixed coefficients is zero, yielding exactly $a_s\nabla_\theta R+ b_s\nabla_\theta F$. Recomputing the coefficients after each student update refreshes their dependence on the current observations.

\paragraph{Exact batch gradient.}
Let $\mathcal S_k$ contain the $N_k$ valid response positions in a fresh student batch, and evaluate the controller at $\theta_k$. With $A_s=\partial z_s/\partial\theta$, define
\begin{equation}
\begin{aligned}
\mathcal L_k(\theta)
 &=\frac{1}{N_k}\sum_{s\in\mathcal S_k}
    \left\langle\sg[W_s(\theta_k)],M_\theta(s)\right\rangle_F,\\
\left.\nabla_\theta\mathcal L_k(\theta)\right|_{\theta_k}
 &=\frac{1}{N_k}\sum_{s\in\mathcal S_k}A_s^\top
    \big(a_sg_{R,s}+b_sg_{F,s}\big)
  =\frac{1}{N_k}\sum_{s\in\mathcal S_k}A_s^\top g_s^*.
\end{aligned}
\label{eq:batch_gradient}
\end{equation}
For plain SGD, $\theta_{k+1}=\theta_k-\eta_k\nabla_\theta\mathcal L_k(\theta_k)$. Other optimizers consume the same gradient with their own parameter-space transformation. The coefficients are deterministic functions of current distributions. Each update refreshes the matrix and coefficient computation, with the complete autoregressive prefix serving as the conditioning state.

\subsection{Complete current-state coefficient rule}
\label{app:controller}

\paragraph{Self-observations to preferences.}
Define
\begin{equation}
\begin{gathered}
 h_S=\frac{M_{11}}{M_{11}+h_0},\qquad
 h_T=\frac{M_{22}}{M_{22}+h_0},\\
 \nu_F=\epsilon+h_T,\qquad
 \nu_R=\epsilon+(1-h_T)h_S,\qquad
 \omega_j=\frac{\nu_j}{\nu_R+\nu_F},\quad j\in\{R,F\}.
\end{gathered}
\label{eq:preferences}
\end{equation}
Here $h_0,\epsilon>0$. The $\nu_j$ are unnormalized preferences; they are distinct from the realized coefficients $a_s,b_s$ below. 

Increasing $h_T$ shifts preference toward the teacher-sourced signal; increasing $h_S$ at fixed $h_T$ shifts it toward the student-sourced signal. A concentrated teacher paired with a diffuse student therefore emphasizes the teacher's assessment of student-distributed candidates. A diffuse teacher gives greater priority to representing its distributed support. The positive floor retains both views when both distributions are concentrated, including when their favored candidates differ. This continuous entropy-aware parameterization instantiates the preference layer of the controller. The compatibility and bargaining results hold for every positive preference pair, separating the structural guarantee from the specific preference map.

\paragraph{Cross-observation excesses to direction.}
Compute $g_R,g_F$ from Eq.~\eqref{eq:gradients}, $n_j=\norm{g_j}$, $e_j=g_j/n_j$, and $c=e_R^\top e_F$. Let $\kappa,D$ be the closed-form quantities in Eq.~\eqref{eq:root}. Then $d^*=(\kappa e_R+e_F)/D$. The ratio $\kappa$ is an algebraic solution of the two-objective bargaining problem; it is distinct from the iteration index $k$.

\paragraph{Discrepancy to position strength.}
With $\tau>0$, set
\begin{equation}
\begin{gathered}
 B=\frac12\sum_i(\sqrt{p_i}-\sqrt{q_i})^2\in[0,1],\\
 w=\arg\max_{0\leq u\leq1}\left\{uB-\frac{\tau}{2}u^2\right\}
  =\clip(B/\tau,0,1).
\end{gathered}
\label{eq:gate}
\end{equation}
This bounded discrepancy is computed directly from the aligned distributions and controls position strength. It complements the matrix's organization of expected scores with a full-distribution measure of agreement. Hellinger-based supervision also appears in sampled excess-token reward shaping \citep{tide}; here it determines the position-level coefficient. Appendix~\ref{app:guarantees} proves its unique scalar optimum.

\paragraph{Realized loss coefficients.}
Expanding $g_s^*=wn_Fd^*$ in the source-centered observation gradients gives
\begin{equation}
 a_s=\frac{wn_F\kappa}{Dn_R},\qquad b_s=\frac{w}{D},\qquad
 g_s^*=a_sg_R+b_sg_F.
\label{eq:coefficients}
\end{equation}
Preferences, realized influence, and total strength have complementary roles: $\omega_j$ values relative gains, the gradient angle determines the compromise, and $wn_F$ sets its norm. Near agreement, the small discrepancy smoothly attenuates the final signal at any preference. The coefficients above make Eq.~\eqref{eq:loss} an exact current-point realization of the controller.

\paragraph{A convenient gradient surrogate.}
A numerically direct realization computes the vector $g_s^*$ and uses
\begin{equation}
 \widetilde{\mathcal L}_s(\theta)
 =\sum_i\sg[g_{s,i}^*]\log q_\theta(i\mid s).
\label{eq:surrogate}
\end{equation}
Since $\one^\top g_s^*=0$, its current logit gradient is exactly $g_s^*$; Appendix~\ref{app:implementation} expands the derivative. The matrix loss and signed surrogate have identical current gradients. Their scalar values follow their respective definitions.

\section{Complete proof of the compatibility identity}
\label{app:geometry}
\begin{lemma}[Logit differentiation and the tangent space]
\label{lem:tangent}
Let $q=\softmax(z)$. Then $\partial q/\partial z=J(q)$, $J(q)\one=0$, and $J(q)$ is positive definite on $\mathcal T=\{v:\one^\top v=0\}$. For any $u,v\in\mathcal T$,
\begin{equation}
 u^\top J(q)^+v=u^\top\operatorname{diag}(q)^{-1}v.
\label{eq:inverselemma}
\end{equation}
\end{lemma}
\begin{proof}
The softmax derivative is $\partial q_i/\partial z_j=q_i(\mathbb{1}_{i=j}-q_j)$. For any $x$, $x^\top Jx=\sum_iq_ix_i^2-(\sum_iq_ix_i)^2=\operatorname{Var}_q(x_i)$. Since every $q_i>0$, this variance is zero exactly when $x$ is constant. Thus the kernel is $\operatorname{span}\{\one\}$ and restriction to $\mathcal T$ is positive definite.

For $v\in\mathcal T$, define $x=\operatorname{diag}(q)^{-1}v$. Then
\[
 Jx=v-q(q^\top\operatorname{diag}(q)^{-1}v)
    =v-q\one^\top v=v.
\]
The vector $J^+v$ is the unique zero-sum solution of this equation; any two solutions differ by a constant vector. Hence $J^+v=x+\alpha\one$ for some scalar $\alpha$. Multiplication by $u^\top$, with $u\in\mathcal T$, proves Eq.~\eqref{eq:inverselemma}. This equality identifies the bilinear forms restricted to the tangent space $\mathcal T$.
\end{proof}

\begin{lemma}[KL gradients and their zero sets]
\label{lem:gradient}
With the teacher frozen, Eq.~\eqref{eq:gradients} holds. Each gradient is zero if and only if $p=q$.
\end{lemma}
\begin{proof}
Let $\ell_i=\log(q_i/p_i)$. Differentiating $R=\sum_iq_i\ell_i$ with respect to $q$ gives $\ell+\one$. Therefore
\[
 \nabla_zR=J(\ell+\one)=q\odot\ell-q(q^\top\ell)=q\odot(\ell-R\one).
\]
For $F=\sum_ip_i\log p_i-\sum_ip_i\log q_i$, the derivative of $\log q_i$ gives $\nabla_zF=q-p$. Each gradient sums to zero. If $g_F=0$, equality of distributions is immediate. If $g_R=0$, positivity of $q_i$ implies $\ell_i=R$ for every $i$, so $q_i=e^Rp_i$. Normalization forces $e^R=1$, and $q=p$. The converse follows by substitution.
\end{proof}

\begin{proof}[Proof of Theorem~\ref{thm:coupling}]
By Lemmas~\ref{lem:tangent} and~\ref{lem:gradient},
\begin{align}
 g_R^\top J^+g_F
 &=\sum_i(q_i-p_i)(\ell_i-R)\\
 &=\sum_iq_i\log(q_i/p_i)+\sum_ip_i\log(p_i/q_i)\\
 &=R+F.
\end{align}
The $R$-multiple disappears because the distributions have equal total mass. Substituting their definitions in Eq.~\eqref{eq:matrix} proves the matrix contrast.

If $p\neq q$, strict nonnegativity of KL implies $R+F>0$, and both gradients are nonzero. Suppose their Euclidean cosine were $-1$. Then $g_R=-\alpha g_F$ for some $\alpha>0$, and
\[
 g_R^\top J^+g_F=-\alpha g_F^\top J^+g_F<0,
\]
contradicting the identity. Thus $c>-1$. The vector
\[
 d_0=\frac{e_R+e_F}{\sqrt{2+2c}}
\]
has unit norm and satisfies $e_R^\top d_0=e_F^\top d_0=(1+c)/\sqrt{2+2c}>0$. This is an explicit common-descent witness. The case $c=1$ is permitted and simply gives $d_0=e_R=e_F$.
\end{proof}

\paragraph{Metric-dependent gradient geometry.}
Equation~\eqref{eq:fisher} weights coordinate products by $1/q_i$, giving small-student-probability coordinates substantial influence. The Euclidean inner product weights these coordinates uniformly. The identity therefore certifies compatibility in the Fisher inverse metric across both acute and obtuse Euclidean configurations. At each interior pair with $p\neq q$, the Euclidean cosine satisfies $c>-1$; the resulting angular margin $1+c$ is pair-dependent.

\section{Extension to strictly convex divergences}
\label{app:general}
The following extension identifies a monotonicity principle underlying the compatibility identity. It establishes a shared descent direction for a finite family of smooth, strictly convex distributional contrasts.

\begin{theorem}[Monotone divergence compatibility]
\label{thm:general}
Let $p,q\in\Delta_V^\circ$, $p\neq q$, and let $f,h:(0,\infty)\to\R$ be continuously differentiable strictly convex functions with $f(1)=h(1)=0$. Define
\[
 D_f(q\Vert p)=\sum_ip_i f(r_i),\quad r_i=q_i/p_i,\quad
 g_f=\nabla_zD_f(q\Vert p),
\]
and similarly define $g_h$. Then
\begin{align}
 g_f &= q\odot\big(f'(r)-\E_q[f'(r)]\one\big),\label{eq:fg}\\
 g_f^\top J^+g_h
 &=\frac12\sum_{i,j}q_iq_j\big(f'(r_i)-f'(r_j)\big)
                    \big(h'(r_i)-h'(r_j)\big)>0.\label{eq:fcov}
\end{align}
For any finite collection $f_1,\ldots,f_m$ satisfying these assumptions, there is a single direction $d$ such that $g_{f_j}^\top d>0$ for every $j$.
\end{theorem}
\begin{proof}
The derivative of $D_f$ with respect to $q_i$ is $f'(r_i)$. Multiplying by the softmax Jacobian gives Eq.~\eqref{eq:fg}. Applying Lemma~\ref{lem:tangent},
\[
 g_f^\top J^+g_h
 =\sum_iq_i(f'(r_i)-\E_qf')(h'(r_i)-\E_qh')
 =\operatorname{Cov}_q(f'(r),h'(r)).
\]
For arbitrary sequences $a_i,b_i$, expansion of the double sum gives
\[
 \operatorname{Cov}_q(a,b)=\tfrac12\sum_{i,j}q_iq_j(a_i-a_j)(b_i-b_j).
\]
The derivative of a differentiable strictly convex function on an interval is strictly increasing: equality of its derivative at two distinct points would force it to be constant between them and the function to be affine on that interval. Hence every summand in Eq.~\eqref{eq:fcov} is nonnegative. Since $p\neq q$, at least two ratios $r_i$ differ, so at least one summand is strictly positive. This proves the pairing claim.

For a finite family, set $v=J^+\sum_{j=1}^m g_{f_j}$. For every $k$,
\[
 g_{f_k}^\top v=\sum_jg_{f_k}^\top J^+g_{f_j}>0.
\]
In particular $v\neq0$, so $d=v/\norm v$ is a unit common-descent direction.
\end{proof}

\paragraph{Generators and assumptions.}
The generators $f(r)=r\log r$ and $h(r)=-\log r$ recover the two source-centered observation contrasts. Adding an affine function to a generator preserves its logit gradient for normalized distributions. Differentiability supplies the gradient representation, and strict convexity supplies the strictly increasing derivatives used in the positivity argument. For general differentiable convex generators, the covariance is nonnegative; strict positivity holds when at least one positively weighted pair contributes a strictly positive product.

\paragraph{Compatibility and the bounded controller.}
The witness $J^+\sum_jg_{f_j}$ establishes feasibility for the finite family. Its norm reflects the inverse metric and can grow near the simplex boundary. \method{} uses a unit Euclidean budget and the explicit two-direction solution to implement bounded local updates. The extension explains the geometric principle, while the source-centered observation construction specifies the candidate populations and statistical meaning of the implemented directions.

\section{Existence, uniqueness, and the closed-form bargaining solution}
\label{app:nash}
We prove Theorem~\ref{thm:nash} and two useful consequences. Let $u_j(d)=e_j^\top d$, $\omega_R+\omega_F=1$, and $\omega_j>0$ throughout this section.

\begin{theorem}[Unique closed-form compromise]
\label{thm:nash}
Under Theorem~\ref{thm:coupling} with $p\neq q$ and $\omega_R,\omega_F>0$, Eq.~\eqref{eq:nash} has a unique solution. Set
\begin{equation}
 \kappa=\frac{c(\omega_R-\omega_F)+\sqrt{c^2(\omega_R-\omega_F)^2+4\omega_R\omega_F}}{2\omega_F},\qquad
 D=\sqrt{\kappa^2+1+2\kappa c}.
\label{eq:root}
\end{equation}
Then $d^*=(\kappa e_R+e_F)/D$, and both $g_j^\top d^*$ are strictly positive.
\end{theorem}

\paragraph{Existence.}
Theorem~\ref{thm:coupling} gives a feasible point $d_0$. On the closed unit ball both utilities lie in $[-1,1]$. For a maximizing sequence with objective at least that of $d_0$, both utilities stay bounded away from zero: approaching zero would send a positive-weight logarithm to $-\infty$, while the other logarithm is at most zero. A convergent subsequence of this sequence therefore has a limit in the strictly positive-utility domain. Continuity proves that the maximum is attained.

\paragraph{The norm constraint is active.}
If a feasible $d$ has $\norm d<1$, then $d/\norm d$ is feasible and increases the objective by $-\log\norm d>0$. Every optimum consequently has unit norm.

\paragraph{Uniqueness.}
Suppose distinct optima $d_1,d_2$ exist. Their midpoint is feasible and, by concavity of the log utilities, has objective at least the optimal value. Strict convexity of the Euclidean ball implies $\norm{(d_1+d_2)/2}<1$ for distinct unit vectors. Radially enlarging the midpoint strictly increases the objective, a contradiction. This proof covers collinear gradients with $c=1$ as well as linearly independent gradients.

\paragraph{First-order conditions.}
The objective is differentiable at the optimum. Its gradient is
\[
 \nabla_d\Phi(d)=\frac{\omega_R e_R}{u_R(d)}+\frac{\omega_F e_F}{u_F(d)}.
\]
Writing the multiplier for $(\norm d^2-1)/2\leq0$ as $\lambda$, the KKT condition is $\nabla\Phi(d)=\lambda d$. Taking the inner product with the unit vector $d$ gives $\lambda=\omega_R+\omega_F=1$. Hence
\begin{equation}
 d=ae_R+be_F,\qquad
 a=\frac{\omega_R}{u_R(d)}>0,\quad
 b=\frac{\omega_F}{u_F(d)}>0.
\label{eq:kktab}
\end{equation}
Here $a,b$ are coefficients of the unit direction in normalized gradients; the state-indexed $a_s,b_s$ in Eq.~\eqref{eq:coefficients} multiply the unnormalized gradients in the final update. It follows that
\begin{equation}
 a^2+cab=\omega_R,\qquad b^2+cab=\omega_F.
\label{eq:ab}
\end{equation}
For $\kappa=a/b>0$, eliminating $b$ yields
\[
 \omega_F\kappa^2+c(\omega_F-\omega_R)\kappa-\omega_R=0.
\]
The constant term is negative and the leading coefficient positive, so exactly one root is positive. This is Eq.~\eqref{eq:root}. Since the actual unique optimum has this positive ratio, normalizing $\kappa e_R+e_F$ reconstructs it. Conversely, this vector satisfies the first-order condition and positive-utility constraints and is therefore optimal by concavity. This completes the proof of Theorem~\ref{thm:nash}.

\begin{corollary}[Scale invariance and proportional fairness]
\label{cor:fair}
The direction $d^*$ is invariant to positive rescaling of either objective at fixed preferences. For every feasible $v$ with $\norm v\leq1$,
\begin{equation}
 \omega_R\frac{u_R(v)-u_R(d^*)}{u_R(d^*)}
 +\omega_F\frac{u_F(v)-u_F(d^*)}{u_F(d^*)}\leq0.
\label{eq:fair}
\end{equation}
\end{corollary}
\begin{proof}
For $\alpha_j>0$, normalization gives $\alpha_jg_j/\norm{\alpha_jg_j}=e_j$, so each utility and its maximizing direction are invariant. By the KKT equation,
\[
 \sum_j\omega_j\frac{u_j(v)}{u_j(d^*)}=d^{*\top}v\leq1.
\]
Subtracting $\sum_j\omega_j=1$ proves Eq.~\eqref{eq:fair}. The inequality characterizes the weighted relative local gains of the two source-centered contrasts.
\end{proof}

\paragraph{Equal weights and the simplest strong baseline.}
If $\omega_R=\omega_F=1/2$, then $\kappa=1$ for every $c>-1$. Thus $d^*=(e_R+e_F)/\norm{e_R+e_F}$. This direction is already a feasible common descent direction, independently of uncertainty. The appropriate incremental question is whether observation-dependent preferences yield more useful learning than this symmetric baseline, at a common norm, gate, learning rate search budget, and teacher cost.
For $c=e_R^\top e_F$, the direct normalized-gradient mixture $v=\alpha e_R+(1-\alpha)e_F$ has unnormalized local gains
\[
 e_R^\top v=\alpha+(1-\alpha)c,\qquad
 e_F^\top v=\alpha c+(1-\alpha).
\]
These expressions characterize how unequal mixing preferences interact with the current gradient geometry.

\paragraph{Stable root evaluation.}
Let $A=c(\omega_R-\omega_F)$ and $Q=\sqrt{A^2+4\omega_R\omega_F}$. An algebraically equivalent computation is
\begin{equation}
 \kappa=\begin{cases}(A+Q)/(2\omega_F),&A\geq0,\\
 2\omega_R/(Q-A),&A<0.
 \end{cases}
\label{eq:stable}
\end{equation}
The second branch avoids subtracting nearly equal positive quantities. Computing $Q$ with a hypot routine further reduces overflow risk. The reference implementation corrects roundoff by clipping the computed cosine to $[-1,1]$ and normalizes the resulting vector directly. Theorem~\ref{thm:coupling} gives $c>-1$ for each distinct interior pair.

\subsection{Aligned-gradient and binary reductions}
\label{app:reductions}
\begin{proposition}[Exact reduction under aligned gradients]
\label{prop:reduction}
If $p\neq q$ and $c=1$, then $g^*=w(q-p)$ for every positive preference pair.
\end{proposition}
\begin{proof}
When $e_R=e_F$, Eq.~\eqref{eq:nash} is maximized by their common unit vector, independently of the preferences. A binary logit tangent space is one-dimensional; Theorem~\ref{thm:coupling} rules out opposite directions, so the normalized gradients coincide.
\end{proof}
Distinct normalized candidate directions create room for preference-dependent coordination. A binary vocabulary yields the exact reduction $g^*=w(q-p)$. The directions also align to first order near agreement, as established in Appendix~\ref{app:asymptotic}. Mismatch and gradient angle together characterize the positions at which coordination changes the signal.

\section{Strength, local descent, and candidate-deficit guarantees}
\label{app:guarantees}
\begin{proposition}[Local descent, boundedness, and teacher-deficit signal]
\label{prop:guarantees}
For $p\neq q$, sufficiently small isolated logit steps $z^+=z-\eta g^*$ strictly decrease both $R$ and $F$. Moreover,
\begin{equation}
 \norm{g^*}=w\norm{q-p}\leq\sqrt{2},\qquad
 -g_j^*\geq w\omega_F(p_j-q_j)\quad\text{whenever }q_j<p_j.
\label{eq:bound}
\end{equation}
For a fixed positive teacher with $p_j=\rho>0$, as $q_j\to0$, Eq.~\eqref{eq:preferences} gives
\begin{equation}
 \liminf_{q_j\to0}(-g_j^*)\ \geq
 \rho\frac{\epsilon}{1+2\epsilon}
 \min\!\left\{\frac{1-\sqrt{1-\rho}}{\tau},1\right\}>0.
\label{eq:tail}
\end{equation}
\end{proposition}
The coordinate lower bound concerns a logit correction. The softmax Jacobian translates this correction into probability changes, whose scale depends on the current student probabilities.

\paragraph{Position-weight optimum.}
The objective $uB-\tau u^2/2$ has second derivative $-\tau<0$, so its constrained maximum is unique. The unconstrained stationary point is $u=B/\tau$; projection onto $[0,1]$ gives Eq.~\eqref{eq:gate}. Because $B>0$ for distinct distributions, $w>0$ whenever $p\neq q$. If $p=q$, the selected weight is zero.

\paragraph{Norm bound.}
For distributions $p,q$,
\[
 \norm{p-q}^2=\sum_i p_i^2+\sum_iq_i^2-2\sum_ip_iq_i\leq2,
\]
since each sum of squares is at most one and the cross term is nonnegative. Since $0\leq w\leq1$ and $\norm{d^*}=1$, Eq.~\eqref{eq:bound} follows. The universal constant is a supremum over pairs approaching distinct point masses. With a bounded parameter-to-logit Jacobian, the corresponding parameter gradient is bounded by the product of its operator norm and this logit bound.

\paragraph{An explicit local descent condition.}
Fix $z$ with $p\neq q$, and let $L_j$ be a Lipschitz constant of $\nabla_z j$ on a ball of radius $r>0$ around $z$, for $j\in\{R,F\}$. Such finite constants exist on every compact logit neighborhood for a fixed positive teacher. Set $m_j=g_j^\top d^*>0$ and $a=wn_F>0$. The descent lemma gives
\begin{equation}
 j(z-\eta ad^*)\leq j(z)-\eta a m_j+\frac{L_j}{2}\eta^2a^2.
\label{eq:descentlemma}
\end{equation}
Thus both losses strictly decrease whenever
\begin{equation}
 0<\eta<\min\!\left\{\frac{r}{a},\frac{2m_R}{L_Ra},\frac{2m_F}{L_Fa}\right\},
\label{eq:etasafe}
\end{equation}
with a term interpreted as $+\infty$ if its Lipschitz constant is zero. The restriction $\eta a<r$ keeps the segment inside the ball. This condition relates the admissible local step to the positive gain margin and the smoothness of each source-centered contrast.

\paragraph{Lower bound for a teacher-supported deficit.}
From Eq.~\eqref{eq:kktab}, $b=\omega_F/(e_F^\top d^*)\geq\omega_F$, because $0<e_F^\top d^*\leq1$. For $q_j<p_j$, $g_{F,j}=q_j-p_j<0$. Also $\log(q_j/p_j)<0$ and $R\geq0$, so $g_{R,j}<0$. The coefficients $a,b$ are positive; hence
\begin{align}
 -g_j^*
 &=-wn_F(ae_{R,j}+be_{F,j})\\
 &\geq-wn_Fbe_{F,j}
 =wb(p_j-q_j)\geq w\omega_F(p_j-q_j).
\end{align}
This argument uses the signs of the individual deficit coordinates and the bargaining KKT equation. The vocabulary construction evaluates every deficit coordinate and supplies the deterministic correction in Eq.~\eqref{eq:bound}.

\paragraph{Uniformly positive source preferences.}
Because $0\leq h_S,h_T<1$,
\[
 \nu_R+\nu_F=2\epsilon+h_T+(1-h_T)h_S\leq1+2\epsilon,
 \qquad \nu_R,\nu_F\geq\epsilon.
\]
Both preferences therefore obey $\omega_j\geq\epsilon/(1+2\epsilon)>0$.

\paragraph{The limit as student support vanishes.}
Fix $p_j=\rho\in(0,1)$. Cauchy--Schwarz on the remaining coordinates implies
\[
 \sum_i\sqrt{p_iq_i}
 \leq\sqrt{\rho q_j}+\sqrt{(1-\rho)(1-q_j)}.
\]
Consequently,
\[
 \liminf_{q_j\to0}B\geq1-\sqrt{1-\rho}>0.
\]
Combining this bound, continuity and monotonicity of $\min(B/\tau,1)$, the preference floor, and $p_j-q_j\to\rho$ gives Eq.~\eqref{eq:tail}. This proves Proposition~\ref{prop:guarantees}.

\begin{proposition}[Finite virtual-step probability-ratio control]
\label{prop:ratios}
Let $q'=\softmax(z-\eta g^*)$, for any $\eta\geq0$. Then
\begin{equation}
 \left|\log\frac{q_i'}{q_i}\right|
 \leq2\eta\lVert g^*\rVert_\infty\leq2\sqrt2\eta
 \quad\text{for every }i.
\label{eq:ratios}
\end{equation}
In particular, both $\KL(q'\Vert q)$ and $\KL(q\Vert q')$ are at most $2\eta\lVert g^*\rVert_\infty$.
\end{proposition}
\begin{proof}
Write $v=-\eta g^*$. Then $\log(q_i'/q_i)=v_i-\log\sum_jq_je^{v_j}$. The logarithm of the weighted exponential average lies between $\min_jv_j$ and $\max_jv_j$. The absolute difference is therefore at most $\max_jv_j-\min_jv_j\leq2\lVert v\rVert_\infty$. Taking expectations under either distribution gives the KL bounds. The bound describes an isolated logit step.
\end{proof}

\section{Near-agreement asymptotics and position strength}
\label{app:asymptotic}
\begin{proposition}[Cubic attenuation]
\label{prop:cubic}
Fix $p\in\Delta_V^\circ$ and $h_0,\epsilon,\tau>0$. Let $q=p+\delta\in\Delta_V^\circ$, $\one^\top\delta=0$, and $\norm\delta\to0$. Then
\begin{equation}
 g^*=\frac{1}{8\tau}\left(\sum_i\frac{\delta_i^2}{p_i}\right)\delta
       +O(\norm\delta^4).
\label{eq:cubic}
\end{equation}
In particular, $\norm{g^*}=\Theta(\norm\delta^3)$ along any nonzero fixed perturbation direction.
\end{proposition}
\begin{proof}
Because $p$ is fixed in the interior, Taylor expansions are uniform in a sufficiently small neighborhood. For each coordinate,
\[
 \sqrt{p_i+\delta_i}-\sqrt{p_i}
 =\frac{\delta_i}{2\sqrt{p_i}}+O(\delta_i^2).
\]
Squaring and summing gives
\begin{equation}
 B=\frac18\sum_i\frac{\delta_i^2}{p_i}+O(\norm\delta^3).
\label{eq:Btaylor}
\end{equation}
For small enough $\delta$, $B<\tau$, so the gate is unsaturated and $w=B/\tau$.

Also, $\log(1+\delta_i/p_i)=\delta_i/p_i+O(\delta_i^2)$ and $R=O(\norm\delta^2)$. Equation~\eqref{eq:gradients} therefore gives
\[
 g_F=\delta,\qquad g_R=\delta+O(\norm\delta^2).
\]
Writing $v=\delta/\norm\delta$ for nonzero $\delta$, it follows that $e_F=v$ and $e_R=v+O(\norm\delta)$. The preferences are continuous near $p$ and uniformly positive. The positive root $\kappa$ is consequently bounded above and away from zero there. Normalization of $\kappa e_R+e_F$ gives $d^*=v+O(\norm\delta)$, and hence
\[
 n_Fd^*=\delta+O(\norm\delta^2).
\]
Multiplying by Eq.~\eqref{eq:Btaylor} divided by $\tau$ yields
\[
 g^*=\frac{1}{8\tau}\left(\sum_i\frac{\delta_i^2}{p_i}\right)\delta
       +O(\norm\delta^4),
\]
as claimed. Positive definiteness of $\operatorname{diag}(p)^{-1}$ supplies the matching upper and lower cubic orders along any fixed nonzero direction.
\end{proof}

\paragraph{Interpretation.}
Near agreement, the two source-centered gradients agree to first order. Preference-dependent changes enter at higher order, and the gate controls the leading attenuation. Equation~\eqref{eq:cubic} quantifies the resulting cubic allocation of supervision to small discrepancies, including its slower fine-matching scale. A positive gate floor would restore a first-order term. The unit-gate ablation isolates the contribution of discrepancy-dependent attenuation.

\paragraph{Behavior at agreement and concentrated predictions.}
For $p=q$, the exact update is zero at every entropy level, preserving the matching distribution. Concentrated predictions with different favored candidates can have large Hellinger discrepancy and receive strong supervision. Position strength thus responds to cross-model mismatch, while self-observations condition the relative candidate-source influence.

\section{Comparison related works}
\label{rel}
\begin{table*}[ht]
\centering
\caption{
Comparison of entropy and token‑selection‑based OPD methods under our
four‑observation framework.
$\bullet$ denotes an observation directly used for token selection or
divergence routing, while $\circ$ denotes an observation involved only
through a derived KL contrast.
}
\label{tab:four_obs_comparison}
\small
\setlength{\tabcolsep}{5pt}
\renewcommand{\arraystretch}{1.18}
\begin{tabularx}{\textwidth}{l *{4}{c}}
\toprule
\textbf{Method}
& $\mathbf{M_{11}}$
& $\mathbf{M_{12}}$
& $\mathbf{M_{21}}$
& $\mathbf{M_{22}}$ \\
& {\scriptsize $H(q)$}
& {\scriptsize $H(p,q)$}
& {\scriptsize $H(q,p)$}
& {\scriptsize $H(p)$} \\
\midrule
REOPOLD  & $\bullet$ & -- & -- & -- \\
TIP      & $\bullet$ & -- & -- & -- \\
EOPD     & -- & -- & -- & $\bullet$ \\
SEAD     & $\bullet$ & -- & -- & $\bullet$ \\
\midrule
\textbf{OC‑OPD (Ours)}
& $\boldsymbol{\bullet}$
& $\boldsymbol{\bullet}$
& $\boldsymbol{\bullet}$
& $\boldsymbol{\bullet}$ \\
\bottomrule
\end{tabularx}
\end{table*}

\section{Implementation details}
\label{app:implementation}
Algorithm~\ref{alg:ocopd} gives the main training loop. The following details specify its sampling, masking, and numerical realization.
\subsection{Training semantics and numerical implementation}
\paragraph{Current-policy sampling.}
The mathematical formulation samples from the full current student distribution $q$ at temperature one. The sampled batch is fixed for one student update, after which fresh rollouts are collected.

\paragraph{Alignment and masking.}
Teacher and student distributions share vocabulary semantics and next-token positions. Both models evaluate the exact student continuation. The loss sums over valid response positions, including the response end token. The signed token-vocabulary surrogate is summed over these positions and divided by their global count $N$. This preserves the position strength in Eq.~\eqref{eq:update}.


\paragraph{Degeneracies and stable normalization.}
Away from agreement, the two gradient norms are nonzero by Theorem~\ref{thm:coupling}. The stable quadratic-root expression in Appendix~\ref{app:nash} supports accurate evaluation under unequal preferences. Direct normalization of the combined vector gives the final gradient at the prescribed magnitude.

\paragraph{Loss realization.}
Because $g^*$ is zero-sum,
\[
 \frac{\partial}{\partial z_j}\sum_i\sg[g_i^*]\log q_i
 =g_j^*-q_j\sum_i g_i^*=g_j^*.
\]
Thus the signed surrogate produces the desired current logit gradient with one student backward pass. The coefficients and controller computations are detached; student log probabilities retain their computation graph. The teacher remains frozen. The frozen-coefficient convention realizes the matrix differential in Eq.~\eqref{eq:gradients} at the current snapshot.


\subsection{Real-update diagnostics}
\label{app:update_diagnostics}

\paragraph{Purpose.}
The geometric analysis in the main text is expressed in terms of a change in
student logits, while optimization acts on shared neural-network parameters. We
therefore measure whether the logit-space quantities predict the
change produced by a real optimizer step. This diagnostic uses a Qwen3-1.7B
non-thinking student and the Qwen3-30B-A3B-Instruct-2507 teacher. Training starts from the same zero-LoRA
initialization used by the corresponding non-thinking experiments.

\paragraph{Quantities measured.}
For a fixed token prefix, let $p=\operatorname{softmax}(z_T)$ be the frozen
teacher distribution and let $q_0=\operatorname{softmax}(z_0)$ denote the
student distribution immediately before the optimizer update. We define
\begin{align}
R(z_0) &= \mathrm{KL}(q_0\,\|\,p),
&F(z_0) &= \mathrm{KL}(p\,\|\,q_0),\\
g_R &= q_0\odot\bigl(\log q_0-\log p-R(z_0)\bigr),
&g_F &= q_0-p.
\end{align}
After the production backward pass and AdamW update, we evaluate the same prefix
again to obtain $z_1$, set $\Delta z=z_1-z_0$, and compare the predictions
\begin{equation}
\widehat{\Delta R}=g_R^\top\Delta z,
\qquad
\widehat{\Delta F}=g_F^\top\Delta z
\end{equation}
with the measured changes
\begin{equation}
\Delta R=R(z_1)-R(z_0),
\qquad
\Delta F=F(z_1)-F(z_0).
\end{equation}
All quantities are computed over the vocabulary. A negative predicted or
measured change denotes descent for the corresponding objective. We report the
Pearson correlation between predicted and measured changes, the fraction of
token positions for which their signs agree, and whether the prediction
correctly identifies simultaneous descent of both objectives.

\paragraph{Probe construction.}
Measurements are taken every ten successful optimizer updates, always using
the same teacher target before and after the update. We use three probe types.
The on-policy probe samples positions from the current rollout.
The fixed in-domain probe reuses fixed prompt--continuation pairs drawn
from the training domain. The \emph{HMMT25 OOD} probe contains four held-out
prompts with fixed short continuations and is used only to diagnose distribution
shift.

\begin{table*}[t]
\centering
\caption{\textbf{Agreement between first-order predictions and measured KL
changes.} Correlations and sign agreement are computed over token positions.
``Both'' is the accuracy of predicting whether RKL and FKL decrease
simultaneously.}
\label{tab:first_order_diagnostics}
\small
\setlength{\tabcolsep}{4.2pt}
\begin{tabular}{lrrrrrrr}
\toprule
Probe & Updates & Tokens & $r_R$ & Sign$_R$ & $r_F$ & Sign$_F$ & Both \\
\midrule
On-policy rollout & 19 & 464 & 0.9977 & 99.57 & 0.9989 & 99.57 & 99.57 \\
Fixed in-domain   & 25 & 300 & 0.9751 & 99.67 & 0.9992 & 98.00 & 99.00 \\
HMMT25 OOD        & 25 & 400 & 0.9987 & 99.00 & 0.9993 & 100.00 & 99.25 \\
\bottomrule
\end{tabular}
\end{table*}

\begin{figure*}[ht]
  \centering
  \includegraphics[width=\textwidth]{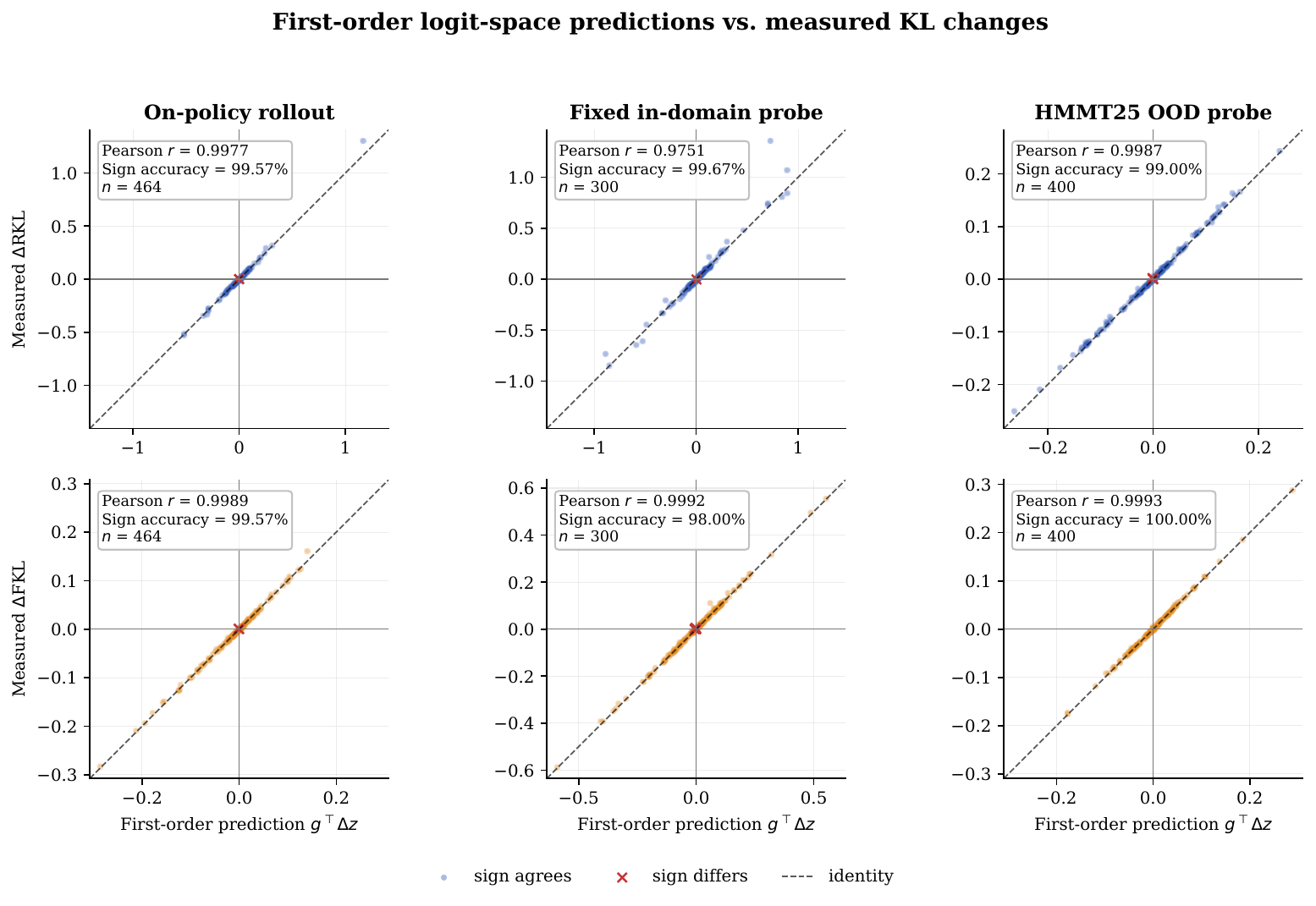}
  \caption{RKL (top) and FKL (bottom) are
  shown separately for on-policy, fixed in-domain, and HMMT25 OOD probes. Each
  point is a token position evaluated immediately before and after the same real
  optimizer step. The dashed line denotes $y=x$ and red crosses indicate sign
  disagreement.}
  \label{fig:first_order_validation_full}
\end{figure*}

Table~\ref{tab:first_order_diagnostics} and
Figure~\ref{fig:first_order_validation_full} show that the local approximation
remains accurate after an update to real network parameters. Pearson
correlations range from 0.9751 to 0.9993, while individual-objective sign
agreement ranges from 98.00\% to 100.00\%. Joint sign accuracy is at least
99.00\% for the two fixed probes and 99.57\% on on-policy rollouts. The few sign
errors occur close to zero, where the measured change is small. These results
empirically validate the local predictive role of $g^\top\Delta z$ under the
observed optimizer steps.

\paragraph{Interpretation and scope.}
The diagnostic supports the specific bridge used by our analysis: for the
optimizer steps observed during training, the logit-space terms
remain accurate predictors of the KL changes induced by a shared-parameter
neural-network update. Agreement on current rollouts, fixed in-domain examples,
and held-out HMMT25 prompts indicates that this fidelity is not confined to one probe distribution.


\section{Experimental details}
\label{app:protocol}
\subsection{Models and experimental organization}
The teacher checkpoints are Qwen3-30B-A3B-Thinking-2507 and Qwen3-30B-A3B-Instruct-2507. The main experiment pairs the Instruct teacher with non-thinking Qwen3-1.7B and Qwen3-4B students. The first auxiliary experiment pairs the Thinking teacher with thinking students at both sizes; the second pairs the Instruct teacher with Qwen3-4B-Base. Within each comparison block, methods use the same student initialization, teacher, prompt source, training budget, generation/evaluation configuration, and metric implementation; only the learning objective or controller component being compared is changed.

Each iteration generates fresh student continuations and aligns teacher and student distributions on the collected prefixes. The four observations, gradient geometry, and coefficients are recomputed at every valid response position. The teacher remains frozen, and the backward pass updates the student. The selected student regime determines the generation configuration of its training and evaluation continuations, and all methods in that regime use the same prompt wrapper and decoding protocol.

\subsection{Training and sampling hyperparameters}
\label{app:hyperparameters}
Table~\ref{tab:hyperparameters} gives the training and sampling settings shared by OPD and OC-OPD within each model family.
\begin{table}[H]
\centering
\caption{Training hyperparameters for the Qwen3 and OLMo experiments. Within each model-family setting, OPD and OC-OPD share the training and sampling hyperparameters. The controller hyperparameters apply only to OC-OPD. Lengths are in tokens. A dash indicates fresh rollouts without reuse.}
\label{tab:hyperparameters}
\begin{tabular}{lcc}
\toprule
Hyperparameter & Qwen3 experiments & OLMo experiments\\
\midrule

\rowcolor{methodshade}
\multicolumn{3}{l}{\textbf{General}}\\
Max prompt length & \multicolumn{2}{c}{8192}\\
GPUs & \multicolumn{2}{c}{$4 \times$ NVIDIA A800 80GB}\\
Training precision & \multicolumn{2}{c}{BF16}\\

\midrule
\rowcolor{methodshade}
\multicolumn{3}{l}{\textbf{Optimization}}\\
Global batch size & 128 & 8\\
Mini-batch size & 32 & 8\\
Mini-batches & 4 & 1\\
Group size & \multicolumn{2}{c}{1}\\
Off-policy epoch & \multicolumn{2}{c}{---}\\
Optimizer & \multicolumn{2}{c}{AdamW}\\
Learning rate & \multicolumn{2}{c}{$1 \times 10^{-5}$}\\
LR schedule & \multicolumn{2}{c}{Constant}\\
Warmup ratio & \multicolumn{2}{c}{0}\\
Weight decay & \multicolumn{2}{c}{0}\\
Adam betas & \multicolumn{2}{c}{$(0.9,0.999)$}\\
Gradient clipping & \multicolumn{2}{c}{Global L2 norm $1.0$}\\

\midrule
\rowcolor{methodshade}
\multicolumn{3}{l}{\textbf{Sampling}}\\
Rollout max response length & \multicolumn{2}{c}{32768}\\
Rollout temperature & \multicolumn{2}{c}{1.0}\\
Rollout top-$p$ & \multicolumn{2}{c}{1.0}\\
Validation max response length & \multicolumn{2}{c}{32768}\\
Validation temperature & \multicolumn{2}{c}{1.0}\\
Validation top-$p$ & \multicolumn{2}{c}{1.0}\\

\midrule
\rowcolor{methodshade}
\multicolumn{3}{l}{\textbf{Method-specific}}\\
$h_0$ & \multicolumn{2}{c}{1.0}\\
$\epsilon$ & \multicolumn{2}{c}{0.05}\\
$\tau$ & \multicolumn{2}{c}{0.01}\\

\bottomrule
\end{tabular}
\end{table}

\subsection{Computational overhead and efficiency}
\label{computational}
We also disentangle and measure the cost of individual components to analyze the extra computational overhead induced by OC-OPD, as outlined below.

We quantify the overhead of OC-OPD relative to standard reverse KL. The table below shows training wall-clock time and step.

\begin{table}[h]
\centering
\caption{Full-run wall-clock time}
\label{tab:walltime}
\begin{tabular}{lcc}
\hline
Method & wall-clock Step & Recorded wall-clock time \\
\hline
OPD / RKL & 250 & 6 h 54 min 29 s \\
OC-OPD & 220 & 6 h 36 min 12 s \\
\hline
\end{tabular}
\end{table}

\textbf{GPU memory.} Full-vocabulary statistics are processed in 128-response-position chunks. For our Qwen3 profiling setup, each position keeps all 151,936 vocabulary entries. One FP32 chunk tensor takes \(74.19\ \text{MiB}\). Chunking restricts intermediate tensor sizes while retaining full vocabulary coverage. On the 10,875-response-token batch, OC-OPD increases the per-GPU PyTorch peak allocated memory by at most 0.435 GiB over standard reverse KL.

\textbf{Parameter-update time.} We profile Qwen3-4B-Base (distilled from Qwen3-30B-A3B-Instruct-2507) on four A800-80GB GPUs over two matched fixed rollout batches (10,875 and 18,043 response tokens). We perform fifteen timed repetitions per batch, then average batch-wise median timings. The measurement includes teacher/student forward passes, backward propagation, synchronization and optimizer step, using precomputed rollouts.
Empirically, OC-OPD adds 0.23 seconds per parameter update, which is approximately a 2.14\% overhead over standard reverse KL. Since the majority of runtime is dominated by student rollout generation, this incremental overhead is modest.

\subsection{Prompt templates}
\label{app:prompts}
The prompt wrapper is fixed by the student generation regime and is shared by all compared methods in that regime. The placeholder \texttt{\{problem\}} is replaced by the benchmark problem text. We use the following templates.

\paragraph{Non-thinking.}
\begingroup
\begin{verbatim}
{problem}\n\nPlease solve the following problem and put your final answer
within
\boxed{}.
\end{verbatim}
\endgroup

\paragraph{Thinking.}
\begingroup
\begin{verbatim}
{problem}\n\nPlease reason step by step, and put your final answer within

\boxed{}.
\end{verbatim}
\endgroup

\paragraph{Base.}
\begingroup
\begin{verbatim}
Problem:\n{problem}
\end{verbatim}
\endgroup

\clearpage
\subsection{OLMo evaluation and checkpoint selection}
\label{app:second-model-family}

We use \texttt{Olmo-3.1-32B-Instruct} as the teacher and \texttt{Olmo-3-7B-Instruct-SFT} as the student. For each of OPD and OC-OPD, we select a checkpoint using the same independent, held-out validation set from DAPO-Math-17k. We evaluate the selected best checkpoints on MATH500 with eight responses per problem and on AIME24 and AIME25 with 32 responses per problem in each repeated decoding evaluation.

\begin{table*}[t]
    \centering
    \caption{OPD and OC-OPD on the OLMo model family. Both methods use the same held-out DAPO-Math-17k validation set for checkpoint selection. AIME24 and AIME25 use repeated decoding evaluations. Scores are percentages. Bold denotes the best result in each column.}
    \small
    \setlength{\tabcolsep}{10pt}
    \renewcommand{\arraystretch}{1.12}
    \begin{tabular}{lcccccc}
        \toprule
        \multirow{2}{*}{\textbf{Method}}
            & \multicolumn{2}{c}{\textbf{MATH500}}
            & \multicolumn{2}{c}{\textbf{AIME 2024}}
            & \multicolumn{2}{c}{\textbf{AIME 2025}} \\
        \cmidrule(lr){2-3}
        \cmidrule(lr){4-5}
        \cmidrule(lr){6-7}
            & Avg@8 & Pass@8
            & Avg@32 & Pass@32
            & Avg@32 & Pass@32 \\
        \midrule
        \rowcolor{gray!12}
        OC-OPD
            & \textbf{81.83} & \textbf{95.40}
            & \textbf{30.04} & \textbf{66.67}
            & \textbf{28.54} & \textbf{60.00} \\
        OPD
            & 69.20 & 92.00
            & 26.45 & 53.33
            & 25.30 & 46.67 \\
        \bottomrule
    \end{tabular}
    \label{tab:olmo-family}
\end{table*}

As shown in Table~\ref{tab:olmo-family}, OC-OPD improves Avg@8
and Pass@8 over OPD on MATH500 by \textbf{12.63} and
\textbf{3.40} percentage points, respectively.
On AIME24, the improvements are \textbf{3.59} percentage points
in Avg@32 and \textbf{13.34} percentage points in Pass@32;
on AIME25, they are \textbf{3.24} and \textbf{13.33}
percentage points, respectively.
These results extend the improvements over OPD to the OLMo
family across all three benchmarks.

\subsection{Controlled objectives and ablations}
\label{app:ablations}
The mathematical ablations use Qwen3-30B-A3B-Instruct-2507 with the Qwen3-4B non-thinking student on DAPO-Math-17k, matching the 4B setting in Table~\ref{tab:main}. All ablations reuse the same initialization, prompts, optimization and sampling setup, and evaluation budget. Table~\ref{tab:math_ablations1} reports the key controls. The comparison design separates the two information paths in Eq.~\eqref{eq:loss}: derivatives of the observations form the candidate-level gradients, while observation values condition their mixture.

\begin{table}[htbp]
\centering
\caption{\textbf{Ablation study on mathematical reasoning.} The student is Qwen3-4B with non-thinking generation and the teacher is Qwen3-30B-A3B-Instruct-2507. AIME24 and AIME25 use repeated decoding evaluations. Scores are percentages; bold denotes the best value in each column.}
\label{tab:math_ablations1}
\small
\setlength{\tabcolsep}{5.0pt}
\renewcommand{\arraystretch}{1.05}
\begin{tabularx}{\linewidth}{>{\raggedright\arraybackslash}p{0.25\linewidth}*{6}{>{\centering\arraybackslash}X}}
\toprule
\multirow{2}{*}{\textbf{Method}}
& \multicolumn{2}{c}{\textbf{AIME 2024}}
& \multicolumn{2}{c}{\textbf{AIME 2025}}
& \multicolumn{2}{c}{\textbf{AIME 2026}} \\
\cmidrule(lr){2-3}\cmidrule(lr){4-5}\cmidrule(lr){6-7}
& Avg@32 & Pass@32 & Avg@32 & Pass@32 & Avg@32 & Pass@32 \\
\midrule
\rowcolor{methodshade}
\method{}
& 60.94 & \textbf{93.33}
& \textbf{56.67} & \textbf{83.33}
& \textbf{58.54} & \textbf{86.67} \\
Forward KL (FKL)
& 55.42 & 83.33
& 53.75 & 73.33
& 54.17 & 83.33 \\

Reverse KL (RKL)
& \textbf{61.04} & 86.67
& 51.04 & 76.67
& 52.19 & 80.00 \\

Adaptive linear mixture
& 60.21 & 86.67
& 53.02 & 80.00
& 53.96 & 80.00 \\

Equal-preference
& 58.65 & 86.67
& 53.54 & 80.00
& 56.67 & 83.33 \\

w/o position gate
& 60.63 & 90.00
& 54.06 & 80.00
& 56.67 & 83.33 \\
\bottomrule
\end{tabularx}
\end{table}

\FloatBarrier

Full \method{} attains the highest Pass@32 on all three AIME editions, while RKL is slightly higher on AIME24 Avg@32. The single-KL controls, adaptive linear mixture, equal-preference variant, and no-gate variant each lose ground on at least one of the coverage metrics. Together, these comparisons support the use of both source-centered directions, observation-conditioned preferences, bargaining-based coordination, and discrepancy-dependent strength.

\paragraph{KL-only controls.}
Forward KL (FKL) optimizes the teacher-sourced contrast $F=\KL(p\Vert q)$, while reverse KL (RKL) optimizes the student-sourced contrast $R=\KL(q\Vert p)$. These rows isolate the two candidate-source objectives without the observation-conditioned coordination used by \method{}.

\paragraph{Adaptive linear mixture}
This variant uses the same observation-conditioned preferences \(\omega_R,\omega_F\), source-centered normalized gradients, and discrepancy gate as OC-OPD, but replaces the bargaining solution with the normalized linear direction \(d_{\mathrm{lin}}=(\omega_Re_R+\omega_Fe_F)/\lVert\omega_Re_R+\omega_Fe_F\rVert_2\). The final magnitude remains \(wn_F\), matching the full method.

\paragraph{Equal preferences.}
Equal-preference bargaining variant fixes $\omega_R=\omega_F=1/2$ while retaining the source-centered gradients, bargaining rule, and discrepancy gate. It therefore removes observation-conditioned preference adaptation without changing the two candidate sources or the update-strength mechanism.

\paragraph{No position gate.}
This variant sets $w=1$, preserving the original preference rule and coordinated direction while removing discrepancy-dependent attenuation of the update magnitude. It directly tests the position-strength component analyzed in Appendix~\ref{app:asymptotic}.


\Needspace{8cm}
\section{Variability across decoding evaluations}
\label{app:aime_stats}

For AIME24 and AIME25, we use repeated decoding evaluations with fixed checkpoints. The comparison in Figure~\ref{fig:variance} uses six decoding evaluations with different sampling seeds for each checkpoint. 

The markers in Figure~\ref{fig:variance} show the reported modal Pass@32 for each setting. This is the most frequently observed solved-problem count across the six decoding evaluations, expressed as a percentage. The whisker endpoints of the error bars correspond to the recorded upper and lower bounds.

\begingroup
\setlength{\intextsep}{4pt}
\captionsetup{skip=4pt}
\begin{figure}[H]
    \centering
    \includegraphics[width=0.9\linewidth]{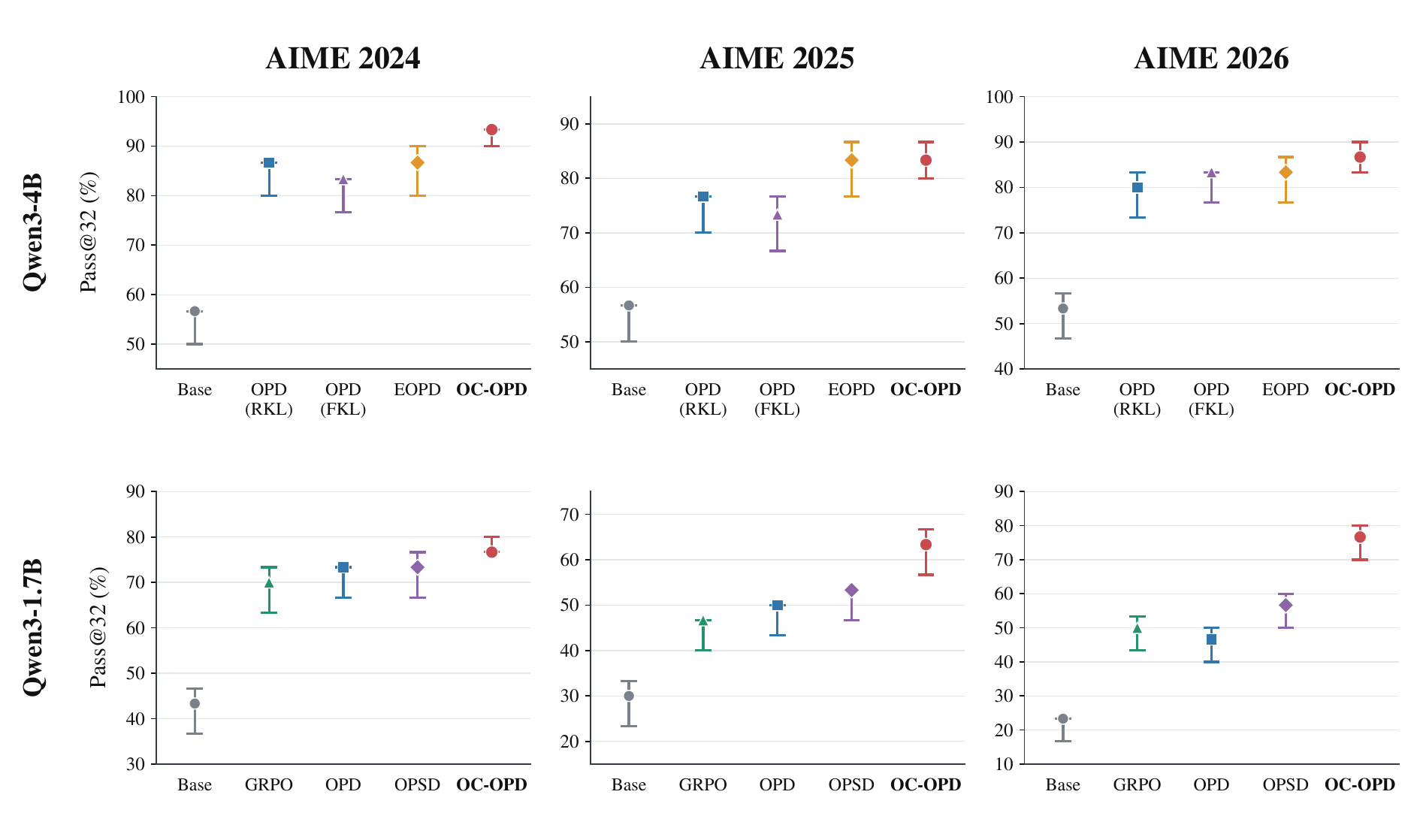}
    \caption{Markers show modal Pass@32 (\%) across six repeated decoding evaluations for fixed Qwen3-1.7B and Qwen3-4B checkpoints. AIME26 is included for comparison.}
    \label{fig:variance}
\end{figure}
\endgroup

\end{document}